\documentclass{article}

\usepackage[final]{neurips_2025}

\usepackage[utf8]{inputenc} 
\usepackage[T1]{fontenc}    
\usepackage{hyperref}       
\usepackage{url}            
\usepackage{booktabs}       
\usepackage{amsfonts}       
\usepackage{nicefrac}       
\usepackage{microtype}      
\usepackage[dvipsnames]{xcolor}         
\usepackage{graphicx}
\usepackage{subfigure}
\usepackage{wrapfig}
\usepackage{tcolorbox}
\usepackage{subcaption}
\usepackage{multirow}

\title{Mind the GAP! The Challenges of Scale in Pixel-based Deep Reinforcement Learning}

\author{%
  Ghada Sokar \\
  Google DeepMind\\
  \texttt{gsokar@google.com} \\
  \And
  Pablo Samuel Castro  \\
  Google DeepMind\\
  \texttt{psc@google.com} \\
}

\begin{document}

\maketitle

\begin{abstract}
Scaling deep reinforcement learning in pixel-based environments presents a significant challenge, often resulting in diminished performance. While recent works have proposed algorithmic and architectural approaches to address this, the underlying cause of the performance drop remains unclear. In this paper, we identify the connection between the output of the encoder (a stack of convolutional layers) and the ensuing dense layers as the main underlying factor limiting scaling capabilities; we denote this connection as the {\bf bottleneck}, and we demonstrate that previous approaches implicitly target this bottleneck. As a result of our analyses, we present Global Average Pooling (GAP) as a simple yet effective way of targeting the bottleneck, thereby avoiding the complexity of earlier approaches.
\end{abstract}

\section{Introduction}
Reinforcement Learning (RL) is widely considered one of the most effective approaches for complex sequential decision-making problems \citep{mnih2015humanlevel,vinyals2019grandmaster,bellemare2020autonomous,degrave2022magnetic,wurman2022outracing}, in particular when combined with deep neural networks (typically referred to as deep RL). In contrast to the so-called ``scaling laws'' observed in supervised learning (where larger networks typically result in improved performance) \citep{kaplan2020scaling}, it is difficult to scale RL networks without sacrificing performance. There has been a recent line of work aimed at developing techniques for effectively scaling value-based networks, such as via the use of mixtures-of-experts \citep{ceron2024mixtures}, network pruning \citep{ceron2024pruned}, tokenization \citep{sokar2025dont}, and regularization \citep{nauman2024bigger}. Most of these techniques tend to focus on structural modifications to standard deep RL networks by leveraging sparse-network training techniques from the supervised learning literature.

\citet{ceron2024mixtures} first demonstrated that na{\" i}vely scaling the penultimate (dense) layer in an RL network results in {\em decreased} performance, and proposed the use of soft mixtures-of-experts \citep[SoftMoEs;][]{puigcerver2024from} to enable improved performance from this form of scaling. \citet{sokar2025dont} argued that the gains from SoftMoEs were mostly due to the use of tokenization. Relatedly, \citet{ceron2024pruned} demonstrated that na{\" i}vely scaling the convolutional layers hurts performance, and showed that incremental parameter pruning yields gains that grow with the size of the original, unpruned, network. While effective, all these methods are non-trivial to implement and can result in increased computational costs.

One unifying aspect of the aforementioned works is that they tend to be most effective on networks that process pixel inputs, such as when training on the Arcade Learning Environment (ALE) \citep{bellemare2013arcade}. These networks are typically divided into an {\em encoder} $\phi$ consisting of a series of convolutional layers,
followed by a series of dense layers $\psi$; thus, for an input $x$, the network output is given by $\psi(\phi(x))$. \citet{ceron2024mixtures} and \citet{sokar2025dont} scaled the first layer of $\psi$ while \citet{ceron2024pruned} scaled all layers in $\phi$. It is worth highlighting that $\psi \circ \phi$ is, in practice, a set of weights connecting the output of $\phi(x)$ with the input layer of $\psi$.

\begin{figure}[!t]
  \centering
  \includegraphics[width=0.85\textwidth]{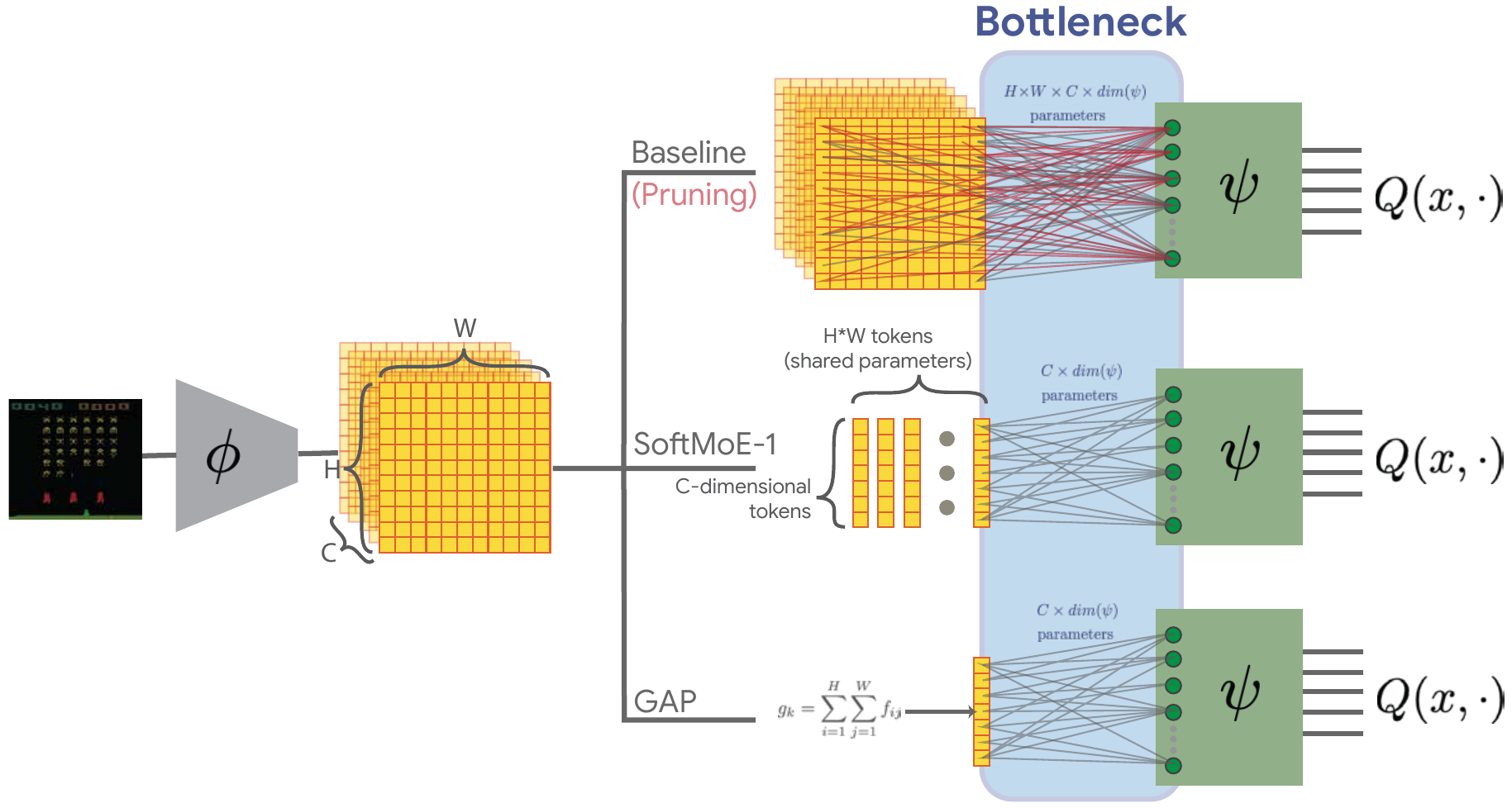}
  \caption{Illustration of the bottleneck in pixel-based networks. Standard dense networks ({\bf Baseline}) connect all $\phi$ outputs with $\psi$, resulting in $H\times W\times C\times dim(\psi)$ parameters (scaled down when using \textbf{pruning}, shown in red). {\bf SoftMoE-1} converts $\phi$'s outputs into $H\times W$ tokens of dimension $C$; the sharing of learned parameters  across tokens results in a bottleneck with $C\times dim(\psi)$ parameters. {\bf GAP} performs average pooling across $H\times W$ spatial dimensions, resulting in $C$ feature maps and $C\times dim(\psi)$ parameters in the bottleneck.}
  \label{fig:rlArchitecture}
\end{figure}

In this work, we argue that the underlying cause behind the effectiveness of the aforementioned methods is that they result in a {\bf bottleneck} between $\phi (x)$ and $\psi$ (see \autoref{fig:rlArchitecture}). As we will argue, performance gains are mostly due to a well-structured bottleneck, rather than the recently proposed architectural modifications to $\psi$. This insight suggests that simpler architectural interventions may be just as effective, which we demonstrate by using Global Average Pooling (GAP) as an effective mechanism for enabling improved performance from scaling.

Our contributions can be summarized as follows:
(i) We investigate the challenges leading to the performance degradation of scaled RL networks; (ii) we study the underlying reasons behind the success of existing architectural approaches in scaling; and (iii) we present pooling as a faster, simpler method yielding superior performance. We begin in Section \ref{sec:preliminaries} by providing the necessary background on reinforcement learning, detailing the architecture employed in pixel-based deep RL, and outlining architectural modifications proposed by recent methods for addressing scaling challenges. Section \ref{sec:experimental_setup} then describes our experimental setup. Our analyses investigating the difficulties of scaling RL networks and how existing methods attempt to address them are presented in Sections \ref{sec:why_scale_fail} and \ref{sec:existing_methods}, respectively. Section \ref{sec:mindtheGAP} is dedicated to our results and analysis on RL networks with GAP. Finally, Section \ref{sec:related_work} covers the related work, followed by a conclusion and discussion in Section \ref{sec:conclusion}. 

\section{Preliminaries}
\label{sec:preliminaries}

\subsection{Reinforcement learning}
Reinforcement learning involves an agent moving through a series of states $x\in\mathcal{X}$ by selecting an action $a\in\mathcal{A}$ at discrete timesteps. After selecting action $a_t$ from state $x_t$, the agent will receive a reward $r_t(x_t, a_t)$, and its goal is to maximize the discounted sum of cumulative rewards $\sum_{t=0}^{\infty}\gamma^t r_t$, where $\gamma\in [0, 1)$ by finding an optimal {\em policy} $\pi:\mathcal{X}\rightarrow \Delta(\mathcal{A})$ which quantifies the agent's behavior at each state. Value-based methods \citep{sutton1998rl} maintain estimates of the value of selecting action $a$ from state $x$ and following $\pi$ afterwards: $Q(x, a) := \mathbb{E}\left[\sum_{t=0}^{\infty}\left[ \gamma^t r_t(x_t, a_t) | x_0 = x, a_0 = a, a_t\sim \pi(x_t)\right]\right] $, where $\pi$ is induced from $Q$, for instance with the use of softmax: $\pi(x)(a) := \frac{e^{Q(x, a)}}{\sum_{a'\in\mathcal{A}}e^{Q(x, a')}}$.

\citet{mnih2015humanlevel} demonstrated deep neural networks can be very effective at approximating $Q$-values, even for complex domains such as Atari games \citep{bellemare2013arcade}; their network has served as the backbone for most deep RL networks. Later, \citet{espeholt2018impala} proposed using ResNet based architecture which demonstrates significant performance improvements over the original convolutional neural network (CNN) architecture. For pixel-based environments, this family of networks consists of a set of convolutional layers \citep{fukushima1980neocognitron}, which we will collectively refer to as $\phi$ (and often referred to as the {\em encoder} or {\em representation}), followed by a set of dense layers, which we will collectively refer to as $\psi$. Thus given an input $x$, the network approximates the $Q$-values as $\tilde{Q}(x,\cdot) = \psi( \phi(x))$.

The output of $\phi$ is a 3-dimensional tensor $H\times W\times C$, where $H$ is height, $W$ is width, and $C$ is the number of feature maps (channels); this output is typically flattened before being fed to $\psi$. Thus, the number of parameters for the functional composition $\psi \circ \phi$ is equal to $H\times W \times C \times dim(\psi)$, where $dim(\psi)$ is the dimensionality of the first dense layer in $\psi$. We refer to connection between $\phi$ and $\psi$ as the {\bf bottleneck}, and it will be the focus of most of our work. See \autoref{fig:rlArchitecture} for an illustration.

\subsection{Network scaling}
As we will demonstrate below, this bottleneck has a direct impact on learning efficiency. The standard approach is to flatten the output of $\phi$ into a single vector before feeding it to $\psi$, which we refer to as an `unstructured' representation. This unstructured representation risks diluting the spatial structure captured by $\phi$, making learning more difficult. This is exacerbated when scaling the width of $\psi$, as demonstrated in prior works \citep{ceron2024mixtures, sokar2025dont}. We hypothesize that the effectiveness of prior architectural modifications for scaling RL networks lies in their ability to induce some form of structure at this bottleneck. The rest of this section details these specific modifications.

\subsubsection{SoftMoE}
To address scaling limitations, \citet{ceron2024mixtures} proposed a notable architectural change: replacing the standard dense layer in $\psi$ with a mixture-of-experts (MoEs). This required first restructuring the 3-dimensional output of the encoder ($\phi$) into a set of tokens to be fed into the mixture. Their investigation into tokenization strategies concluded that forming $H \times W$ tokens, each with $C$ dimensions (the channel depth), yielded the best performance. This restructuring effectively changes the connection, reducing the bottleneck's parameter count to $C \times dim(\psi)$ (See \autoref{fig:rlArchitecture}). This architectural change has proven highly effective for various agents and at different scales.

Building on this, \citet{sokar2025dont} isolated the source of these performance gains. They demonstrated that the tokenization step itself—not the MoE architecture—was the most critical component. Their evidence showed that a single expert model (SoftMoE-1), which retains the tokenization but omits the expert routing, achieves performance nearly identical to that of the full multi-expert model across different scales.

\subsubsection{Sparse networks}
Sparse methods offer an alternative approach to managing network complexity. \citet{sokar2022dynamic} demonstrated that using sparse neural networks in place of dense ones can increase learning speed and lead to improved performance. Following this, \citet{graesser2022state} studied various ways to induce this sparsity, such as by pruning network weights during training or by using networks that are sparse from scratch. By imposing sparsity, these techniques directly structure the bottleneck, reducing the density of its connections. With a given sparsity level $s$, the effective number of parameters in the bottleneck is reduced to $s \times H \times W \times C \times dim(\psi)$. These sparse approaches have shown promise for the scaling of larger architectures; indeed, \citet{ceron2024pruned} recently demonstrated that pruning is an effective approach that facilitates the scaling of networks. 

\paragraph{Gradual pruning} This strategy begins with a standard, fully-dense network. As training progresses, connections (parameters) with low magnitudes—which are considered less salient to the network's function—are progressively removed. This pruning process often follows a polynomial schedule \citep{zhu2017prune}. Once the pruning schedule concludes, the network achieves its target sparsity level, and this fixed sparse architecture is maintained for the remainder of training.

\paragraph{Sparse from scratch} In contrast to gradual pruning, this approach defines a sparse network at initialization, and this specific sparsity level is maintained throughout training. The sparse topology can be \textit{static}, meaning the set of active connections is fixed for the entire training duration. Alternatively, the topology can be \textit{dynamic}. For example, the \textit{RigL} method \citep{evci2020rigging} dynamically optimizes the connections by periodically pruning a portion of the weights and growing new ones elsewhere, effectively "rewiring" the network while maintaining the overall sparsity.

\section{Analyses}
\label{sec:analyses}

\begin{tcolorbox}[colback=blue!5, colframe=blue!50, title=\textbf{Main hypothesis}]
A low-density and well-structured {\bf bottleneck} enables scaling deep RL networks.
\end{tcolorbox}

We conduct a series of analyses, both quantitative and qualitative, to provide evidence for our main hypothesis. We investigate impacts on performance, plasticity, and properties of the learned features. Given the effectiveness of SoftMoEs~\citep{ceron2024mixtures}, Pruning~\citep{ceron2024pruned}, and Tokenization~\citep{sokar2025dont} for scaling value-based networks, our analyses in this section will focus on these.

\subsection{Experimental setup}
\label{sec:experimental_setup}
\paragraph{Architectures} We employ the Impala architecture \citep{espeholt2018impala} for its superior performance, and while our analysis centers on it, we also confirm our findings' broad applicability using the standard CNN architecture \citep{mnih2015humanlevel}. Our main experiments and analyses present results across various scaling factors for the width of $\psi$ (specifically, $\times1$, $\times2$, $\times4$, and $\times8$). When a scaling factor is not explicitly mentioned, our analysis defaults to the $\times4$ scale. This is because, as shown in prior work \citep{ceron2024mixtures, sokar2025dont, ceron2024pruned}, the $\times4$ scale provides a substantial performance improvement, while further scaling tends to yield only marginal gains. Although layer width is our primary focus, we also include some preliminary investigations into scaling network depth in the appendix. 

\paragraph{Agents and environments} Our primary experimental setup involves the Rainbow agent \citep{hessel2018rainbow} evaluated on the Arcade Learning Environment (ALE) suite \citep{bellemare2013arcade}. For direct comparison with recent work, we use the same $20$-game subset from \citet{ceron2024mixtures} and \citet{sokar2025dont}. However, we present our main results on the full ALE suite of 60 Atari games. We ran each experiment for a total of 200 million environment steps, with results averaged over $5$ independent seeds, except for the experiments with an increased replay ratio of 2, where we report the results at 50M steps. To further test the generality of our approach on discrete tasks, we also evaluate Rainbow on the Procgen benchmark \citep{cobbe2019procgen}. Additionally, we assess performance on the data-efficient 100k benchmark \citep{Kaiser2020Model}, for which we use DER, a version of Rainbow specifically tuned for that data-constrained setting. Finally, to demonstrate our findings extend beyond discrete action spaces, we also evaluate the Soft Actor-Critic (SAC) agent \citep{haarnoja2018soft} on continuous control tasks from the DeepMind Control (DMC) suite \citep{tassa2018deepmind}.

\paragraph{Code and compute resources} For all our experiments, we use the Dopamine library\footnote{Dopamine: https://github.com/google/dopamine} with Jax implementations \citep{castro2018dopamine}. For SoftMoE \citep{ceron2024mixtures}, we use the official implementation integrated in Dopamine. For the sparse methods \citep{ceron2024pruned, graesser2022state}, we use the same JaxPruner library\footnote{JaxPruner: https://github.com/google-research/jaxpruner} \citep{lee2024jaxpruner} used by \citep{ceron2024pruned}. All libraries have Apache-2.0 license. All experiments were run on NVIDIA Tesla P100 GPUs. The duration of each experiment ranged from 4 to 13 days, depending on the specific scale and algorithm. We present the exact run time for each case in Section \ref{sec:mindtheGAP}.

\begin{figure}[t]
    \centering
 \includegraphics[width=0.3\textwidth]{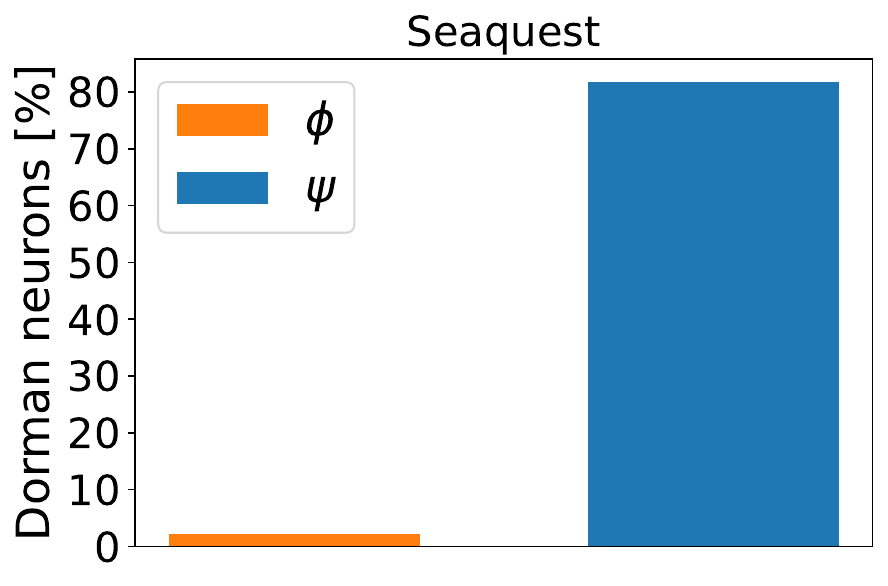}
 \includegraphics[width=0.3\textwidth]{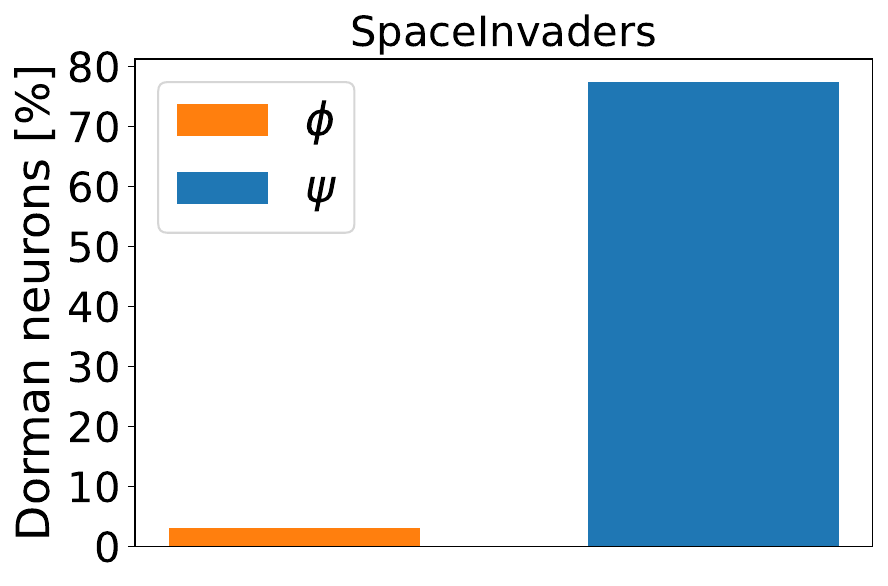}
\hfill
 \includegraphics[width=0.3\textwidth]{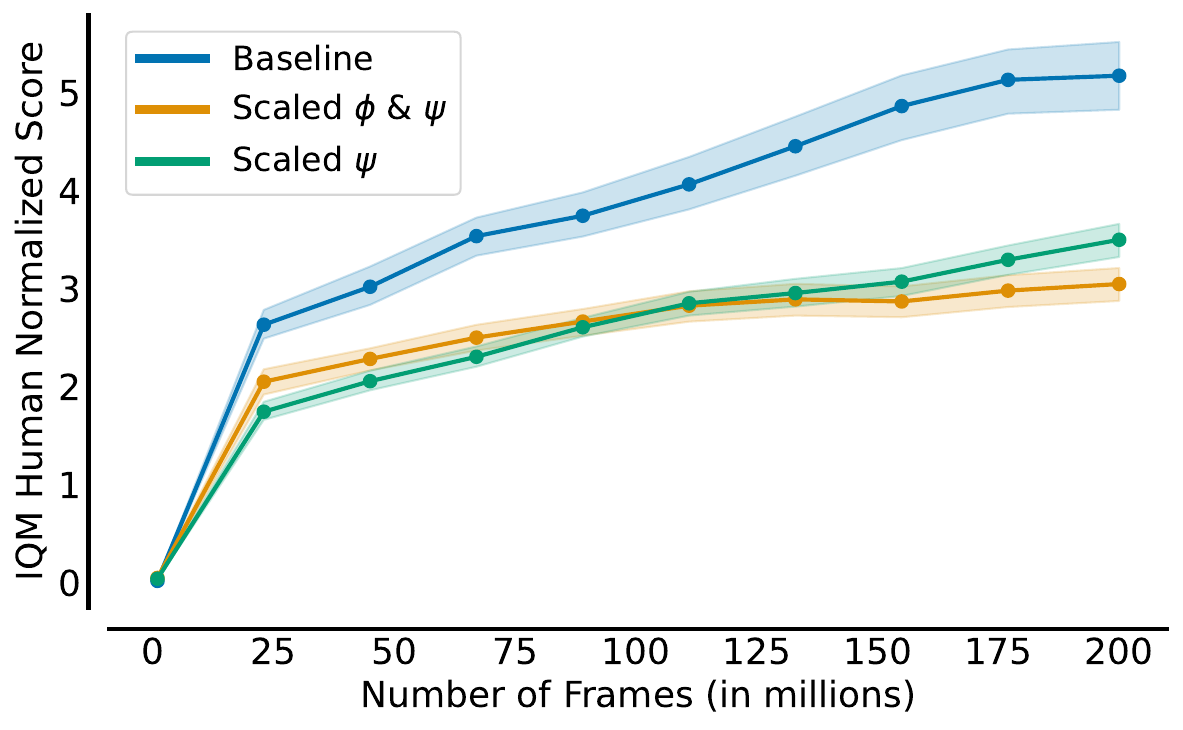}

    \caption{{\bf (Left)} Distribution of dormant neurons across $\phi$ and $\psi$ in scaled baseline across different games at the end of training. The fully connected layer exhibits the highest percentage of dormancy. {\bf (Right)} The performance degradation associated with scaling the entire network architecture is comparable to that observed when only the bottleneck is scaled. The performance is aggregated over 20 games.}
    \label{fig:scalingAllVOneAndStats}
\end{figure}

\paragraph{Implementation details} For all algorithms, we use the default hyperparameters in the Dopamine library. For sparse-training algorithms, we follow \citep{graesser2022state} and use 90\% sparsity. Additional analysis covering other sparsity levels is included in the appendix. For gradual pruning, we start pruning at 8M environment steps and stop at 160M (80\% into training), following the schedules recommended by \citet{graesser2022state} and \citet{ceron2024pruned}. For RigL \citep{evci2020rigging}, we use a drop fraction of 20\% and a connection update interval of 5000. For SoftMoE, we use the single-expert variant (SoftMoE-1). This choice was twofold: first, it has been shown to yield performance comparable to the standard multi-expert SoftMoE \citep{sokar2025dont}, and second, it ensures a more direct architectural comparison with the other methods considered in this study. We report interquartile mean (IQM) with 95\% stratified bootstrap confidence intervals as recommended in \citep{agarwal2021deep}. Full experimental details are provided in the appendix.

\subsection{Why scaling deep RL networks hurts performance}
\label{sec:why_scale_fail}
As has been previously demonstrated,  na{\" i}vely scaling networks deteriorates performance \citep{ceron2024mixtures,ceron2024pruned,nauman2024bigger}. In this section we conduct a series of experiments to diagnose the underlying causes for this difficulty.

We start by analyzing the training dynamics of a network where the width of all layers in both $\phi$ and $\psi$ are uniformly scaled by a factor of $4$. We  examine neuron activity by measuring the fraction of dormant neurons, a metric that serves as a key indicator of network plasticity. Following \citet{sokar2023dormant}, a neuron is considered "dormant" if its average activation falls below a certain threshold. As shown in the left plot of \autoref{fig:scalingAllVOneAndStats}, we find that $\psi$ exhibits a large fraction of dormant neurons, while $\phi$ has low dormancy rates. This suggests that {\bf scaling mostly affects the plasticity of the bottleneck}.   

\begin{figure}[t]
  \centering
  \includegraphics[width=0.205\textwidth]{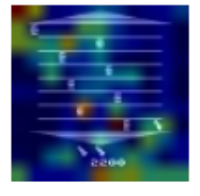}
    \includegraphics[width=0.22\textwidth]{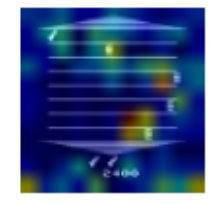}
    \hspace{1cm}
\includegraphics[width=0.2\textwidth]{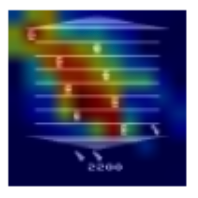}    \includegraphics[width=0.2\textwidth]{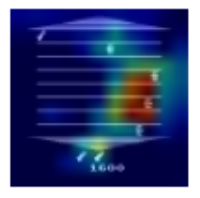}

  \caption{{\bf GAP helps improve attention to relevant areas of input.} Visualizing influential regions for network decisions using Grad-CAM \citep{selvaraju2017grad}. \textbf{(Left)} The scaled baseline fails to attend to the important regions, focusing on irrelevant background details. \textbf{(Right)} GAP attends to the important regions in the input.}
  \label{fig:Saliencymap}
\end{figure}
\begin{figure}[t]
  \centering
  \includegraphics[width=\textwidth]{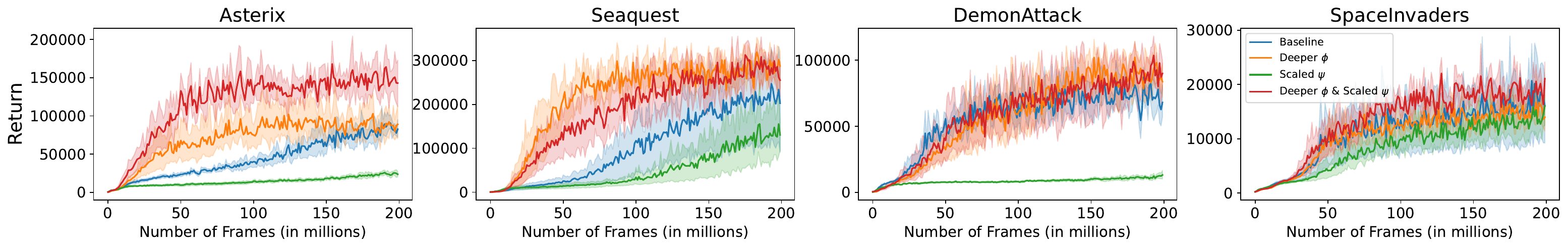}
  \caption{Scaling $\psi$ hinders learning effective combinations of encoder's features, leading to significant performance drop. However, performance dramatically improves when the scaled $\psi$ is fed with higher-level, more abstract features obtain by increasing the depth of $\phi$.}
  \label{fig:feature_learning}
\end{figure}

In the right plot of \autoref{fig:scalingAllVOneAndStats}, we compare the performance of the \textcolor{blue}{baseline network} against the same network with \textcolor{orange}{scaled $\phi$ and $\psi$}. As consistently observed in previous studies, the scaled network exhibits a degradation in performance. To investigate the contribution of $\psi$ in this performance decrease, we only \textcolor{ForestGreen}{scale the bottleneck} (via the first layer 
 of $\psi$) to match the parameter count of the \textcolor{orange}{fully scaled network}. As can be observed, the two scaled models has comparable performance, suggesting that {\bf the bottleneck drives most of the performance degradation when scaling}. 

To interpret the quality of the learned features of the scaled network, we generate saliency maps for the areas that have the greatest impact on the scaled network's output using Grad-CAM \citep{selvaraju2017grad}. \autoref{fig:Saliencymap} (left) reveals that na{\" i}vely scaled networks fail to focus on important regions and focus on irrelevant background areas. This suggests that {\bf scaling the bottleneck impairs a network's ability to process and learn effective combinations of the representation $\phi$}.

To further investigate potential challenges in feature learning within $\psi$, we study the network behavior when providing more abstract, higher-level features to $\psi$. Specifically, we increased the depth of $\phi$ by adding four additional ResNet blocks, while using the scaled bottleneck. This modification makes $\psi$ receive more structured, high-level features from the encoder. We present the performance throughout training in \autoref{fig:feature_learning}. The fact that we observe a dramatic increase in performance when compared to the network that only scaled $\psi$ suggests that \textbf{structured representations helps feature learning in scaled networks.}

\begin{wrapfigure}{r}{0.49\textwidth}
  \centering
  \includegraphics[width=0.435\textwidth]{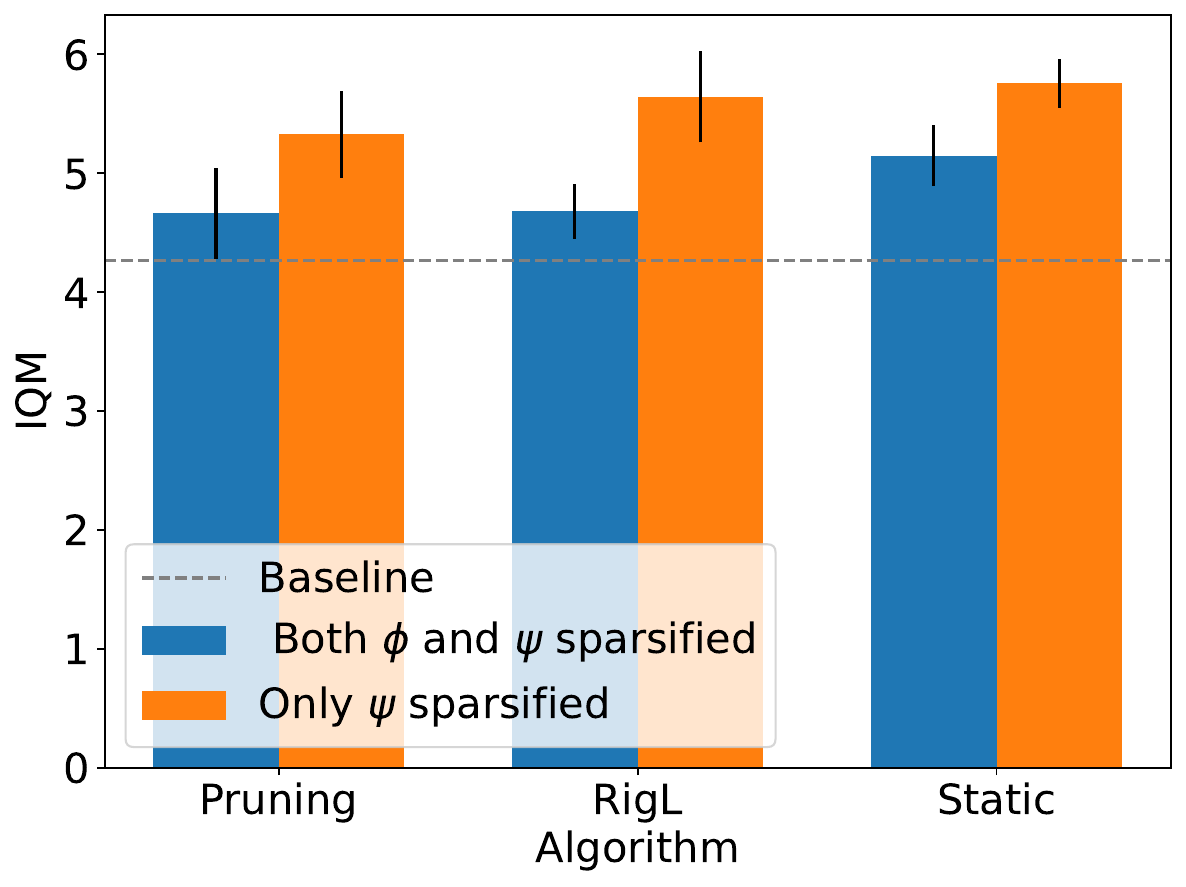}
  
    \includegraphics[width=0.435\textwidth]{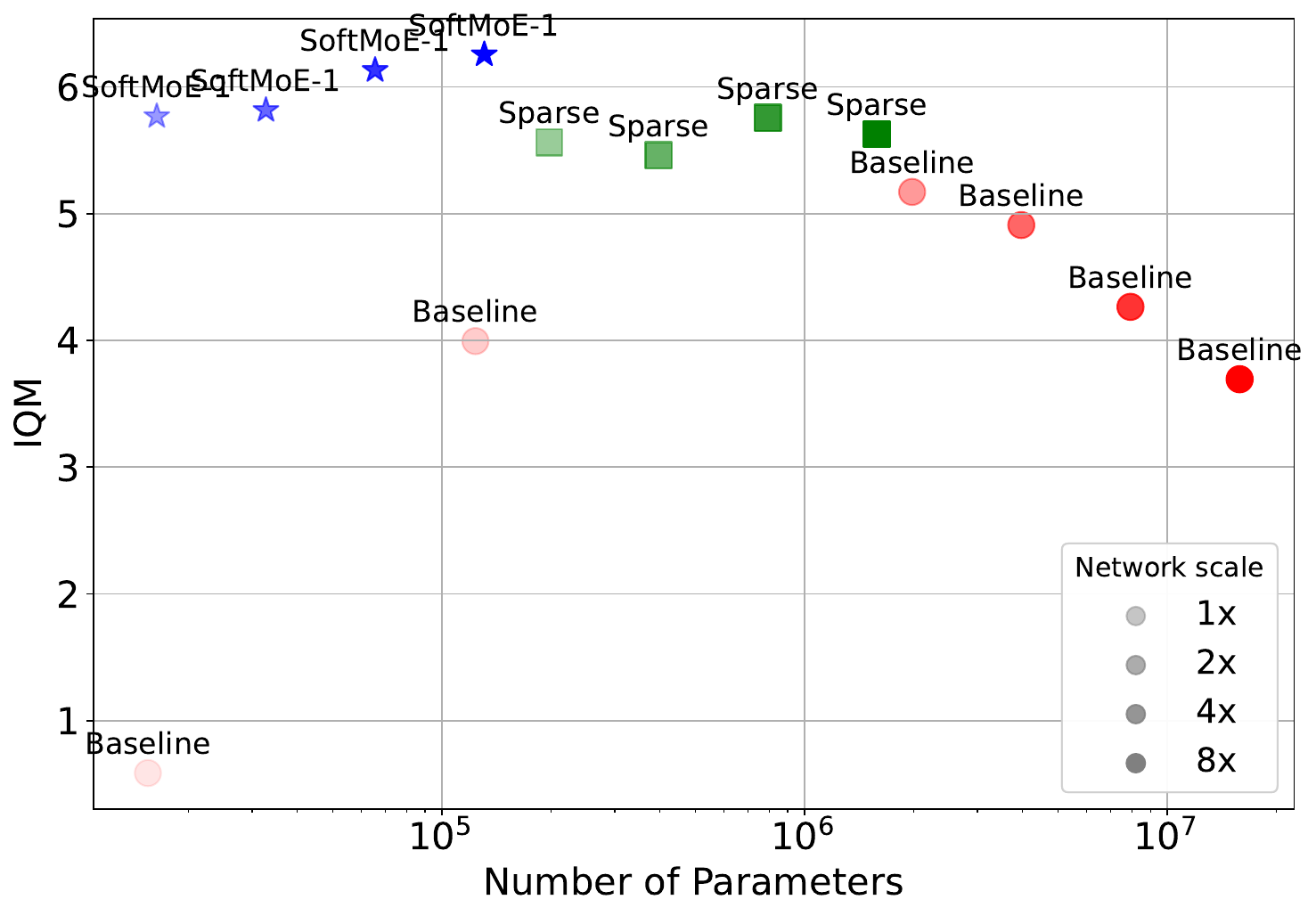}
  \caption{\textbf{(Top)} Across different sparse algorithms, sparsification of only $\psi$ yields better performance than sparsifying $\phi$ and $\psi$. \textbf{(Bottom)} The relation between performance and the effective number of parameters in $\psi$ for different approaches. Architectural methods have lower effective density than the baseline which correlates with the observed performance improvements.}
  \label{fig:sparseallvsone_pensize}
  \vspace{-1em}
\end{wrapfigure}

\subsection{Existing techniques are mainly targetting the bottleneck}
\label{sec:existing_methods}
As previously mentioned, there have been a number of recent proposals to enable scaling deep RL networks by using sophisticated architectural modifications. We hypothesize that a core reason for their effectiveness is that they are implicitly targetting the bottleneck. The strong performance of SoftMoE-1 and tokenized baselines presented by \citet{sokar2025dont} support this claim, as they are primarily re-structuring the encoder output. 

To validate our hypothesis on sparse methods, we evaluated the impact of sparse training techniques when limited to the bottleneck, while keeping all other layers dense. We performed this analysis on gradual pruning \citep{ceron2024pruned}, dynamic sparsity (RigL) and static sparsity \citep{graesser2022state,sokar2022dynamic}. As shown in the top plot of \autoref{fig:sparseallvsone_pensize}, restricting sparsification to the scaled bottleneck results in improved performance across all sparse training techniques. This suggests that {\bf applying sparsity only to the bottleneck is sufficient to enable scaling RL networks}.

In addition to structuring the encoder output, existing techniques effectively reduce the density of the bottleneck by either structuring the output as tokens (as in SoftMoE) which results in a lower dimensional input to $\psi$, or explicitly masking the majority of input weights (as in sparse methods). The bottom plot of \autoref{fig:sparseallvsone_pensize} confirms this reduction in parameters, and illustrates a positive correlation between scale and performance, which is notably absent in the baseline. This suggests that {\bf low-density and structured bottlenecks facilitate scaling deep RL networks}.

\section{Mind the GAP!}
\label{sec:mindtheGAP}
\begin{figure}[t]
  \centering
  \includegraphics[width=0.23\textwidth]{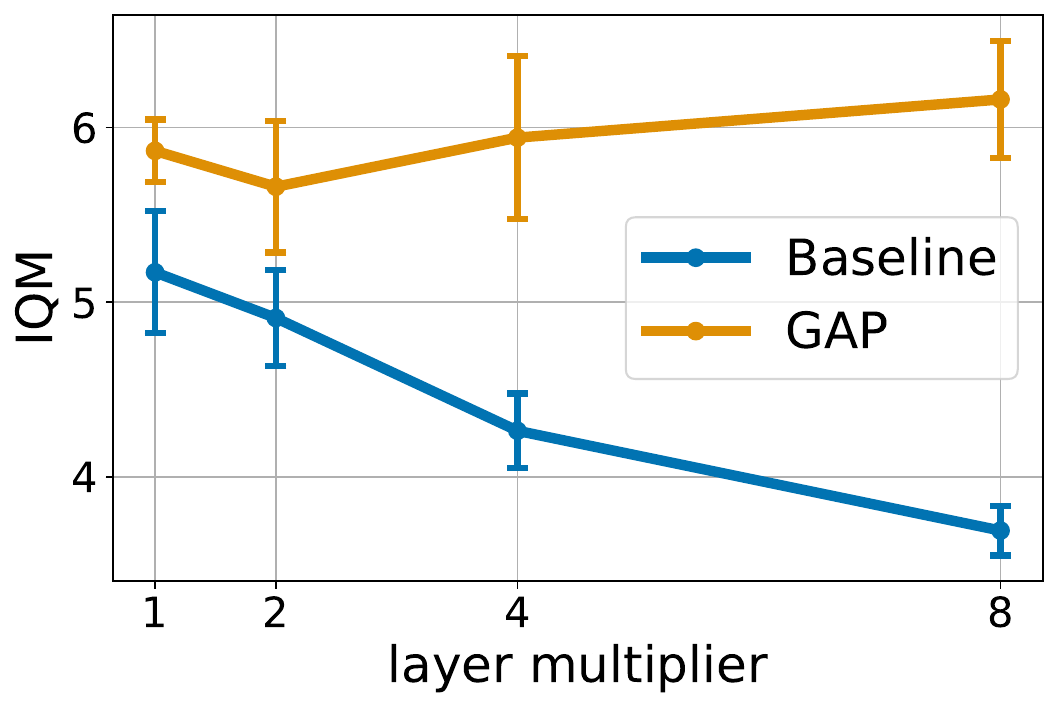}
  \hfill
  \includegraphics[width=0.23\textwidth]{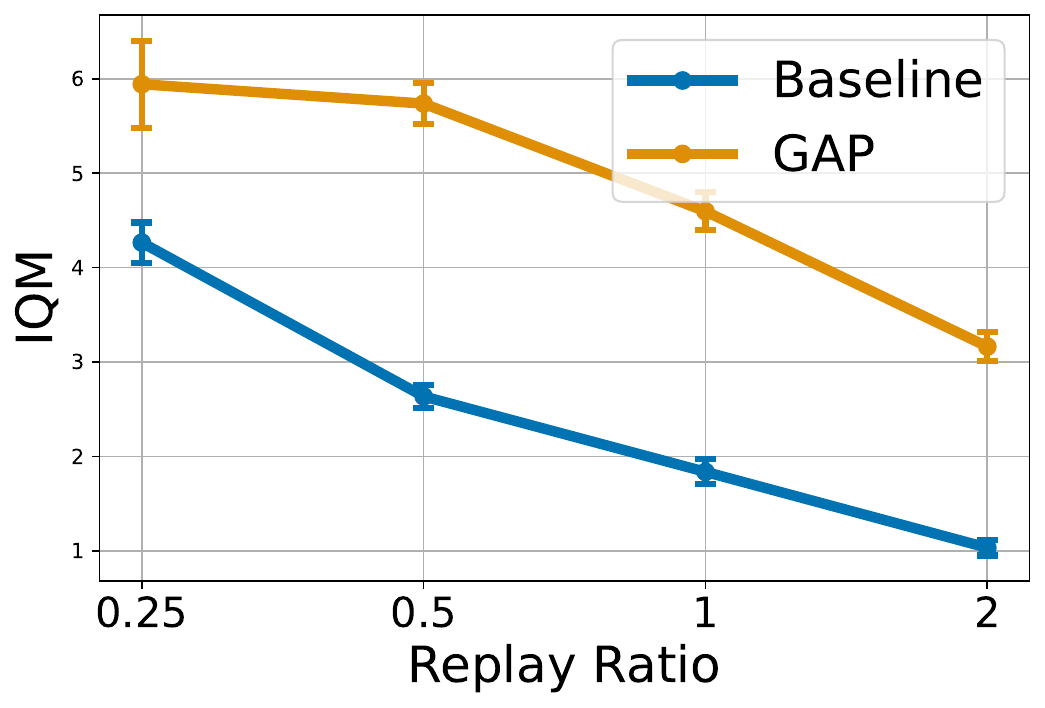}
  \hfill
  \includegraphics[width=0.24\textwidth]{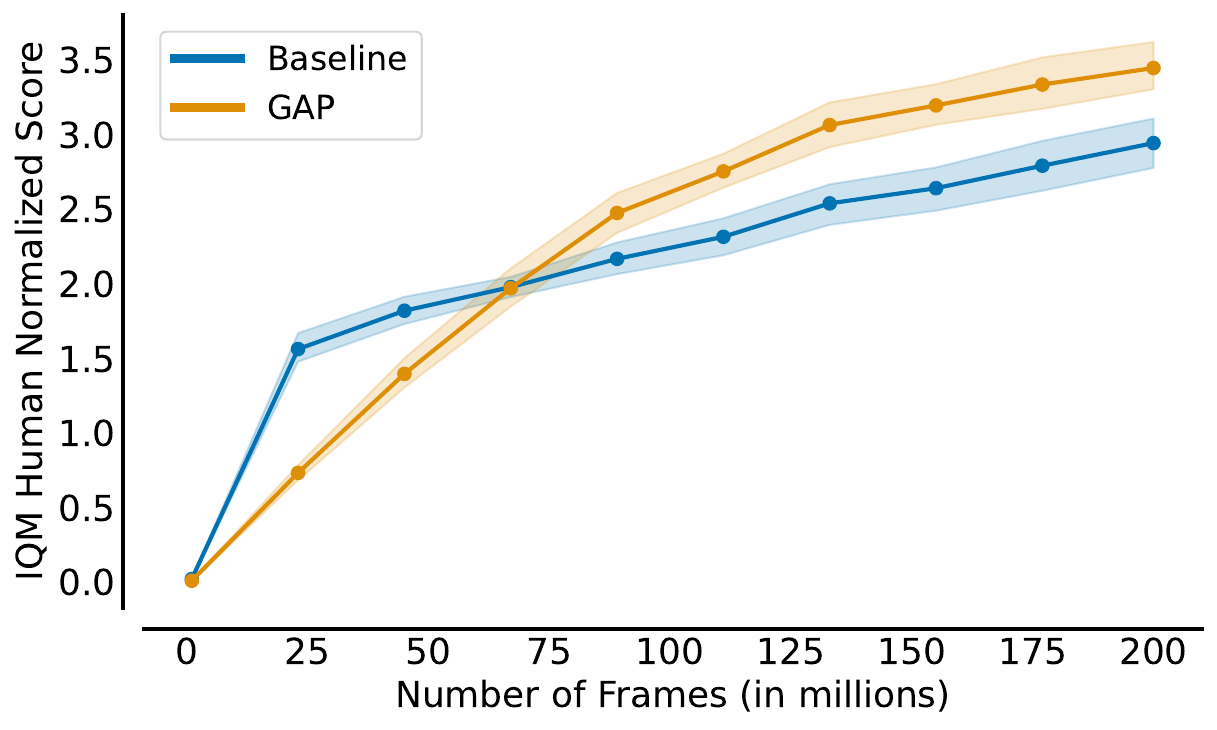}
  \hfill
  \includegraphics[width=0.24\textwidth]{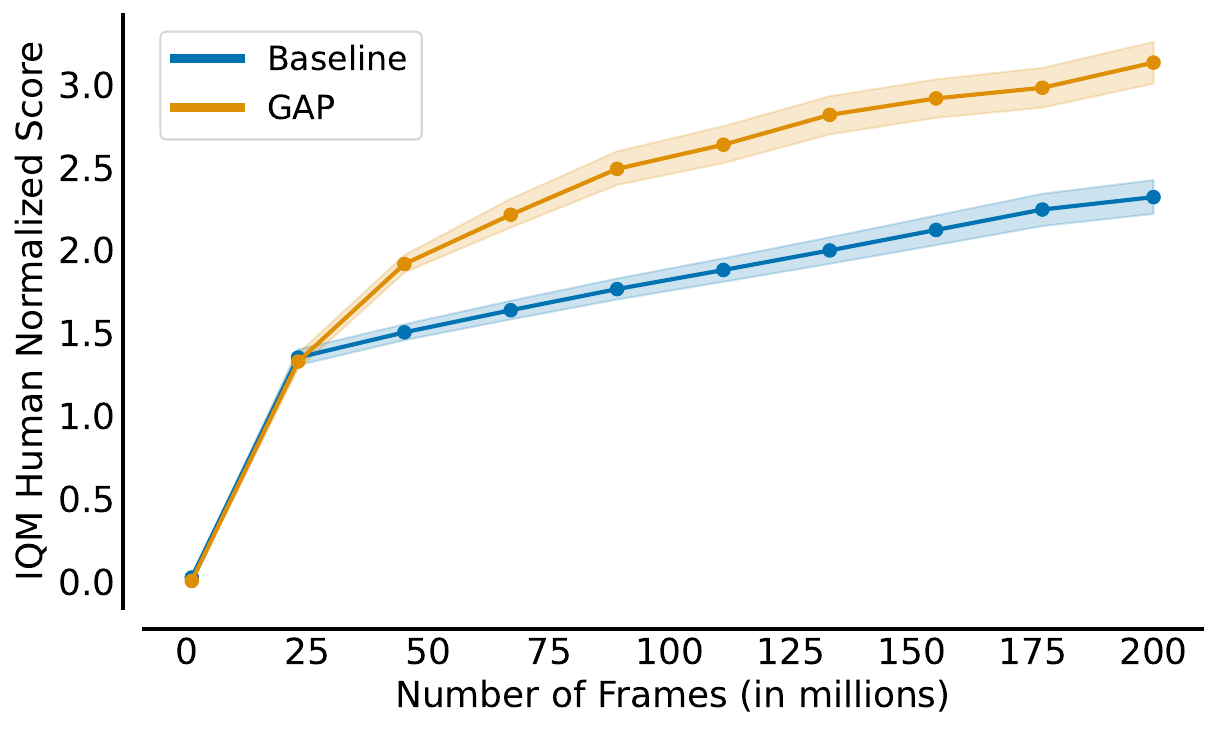}
  \caption{The impact of GAP across wide range of setting. From left to right: performance across different network scales, sample-efficient training with various high replay ratio values (default $= 0.25$), performance of the CNN architecture used by \citep{mnih2015humanlevel}, and performance on the full 60 games of Atari 2600. In all cases, GAP significantly improves performance.}
  \label{fig:GAP}
\end{figure}

\begin{wrapfigure}{r}{0.49\textwidth}
  \centering
  \includegraphics[width=0.49\textwidth]{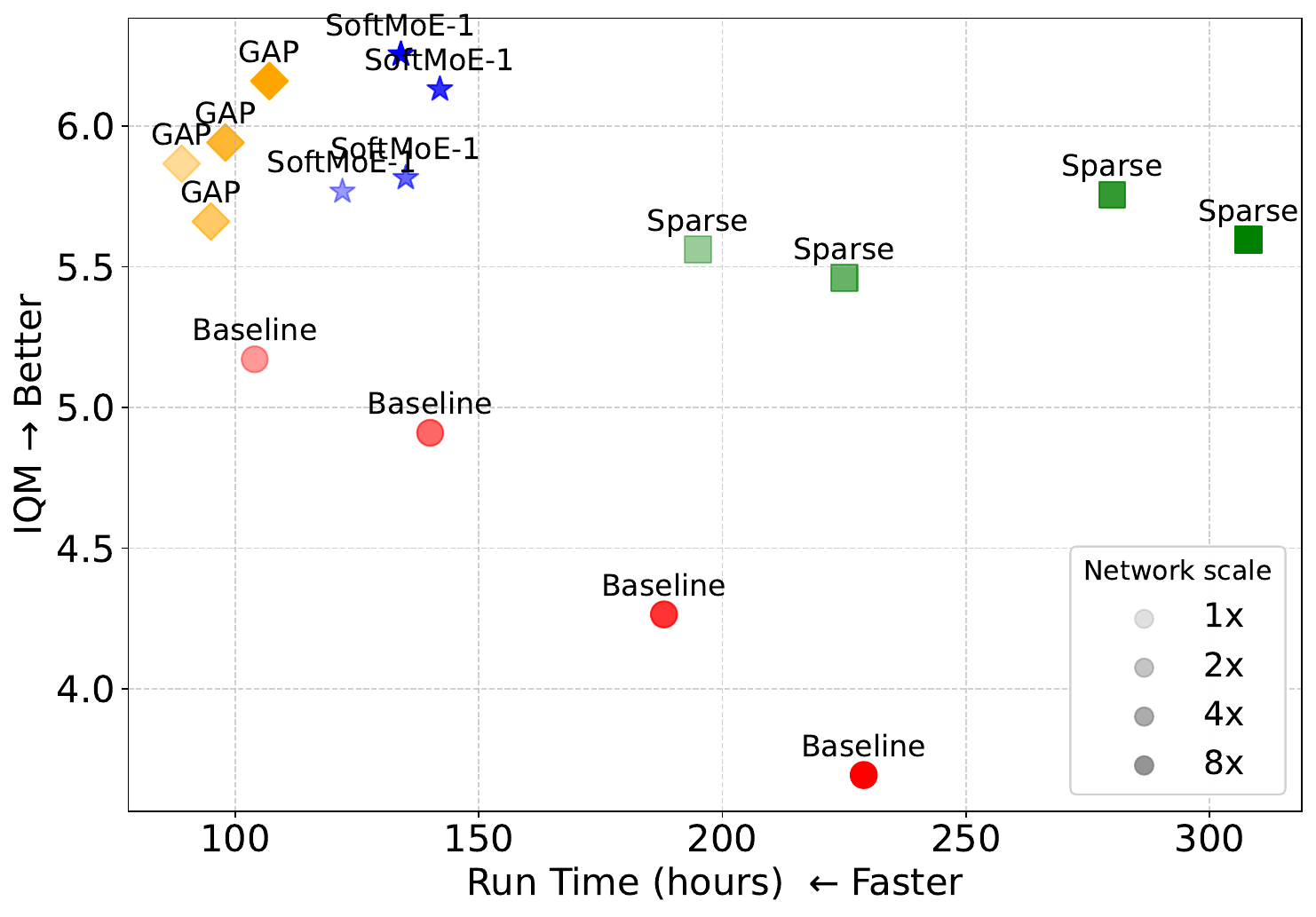}
  \caption{The computational cost versus performance across varying network scales for different algorithms. GAP offers the \textit{highest} speed while obtaining substantial performance improvements.}
  \label{fig:performancevscomputation}
  \vspace{-1em}
\end{wrapfigure}

\begin{figure}[t]
  \centering
    \begin{minipage}{0.48\linewidth}
        \centering
\includegraphics[width=\columnwidth]{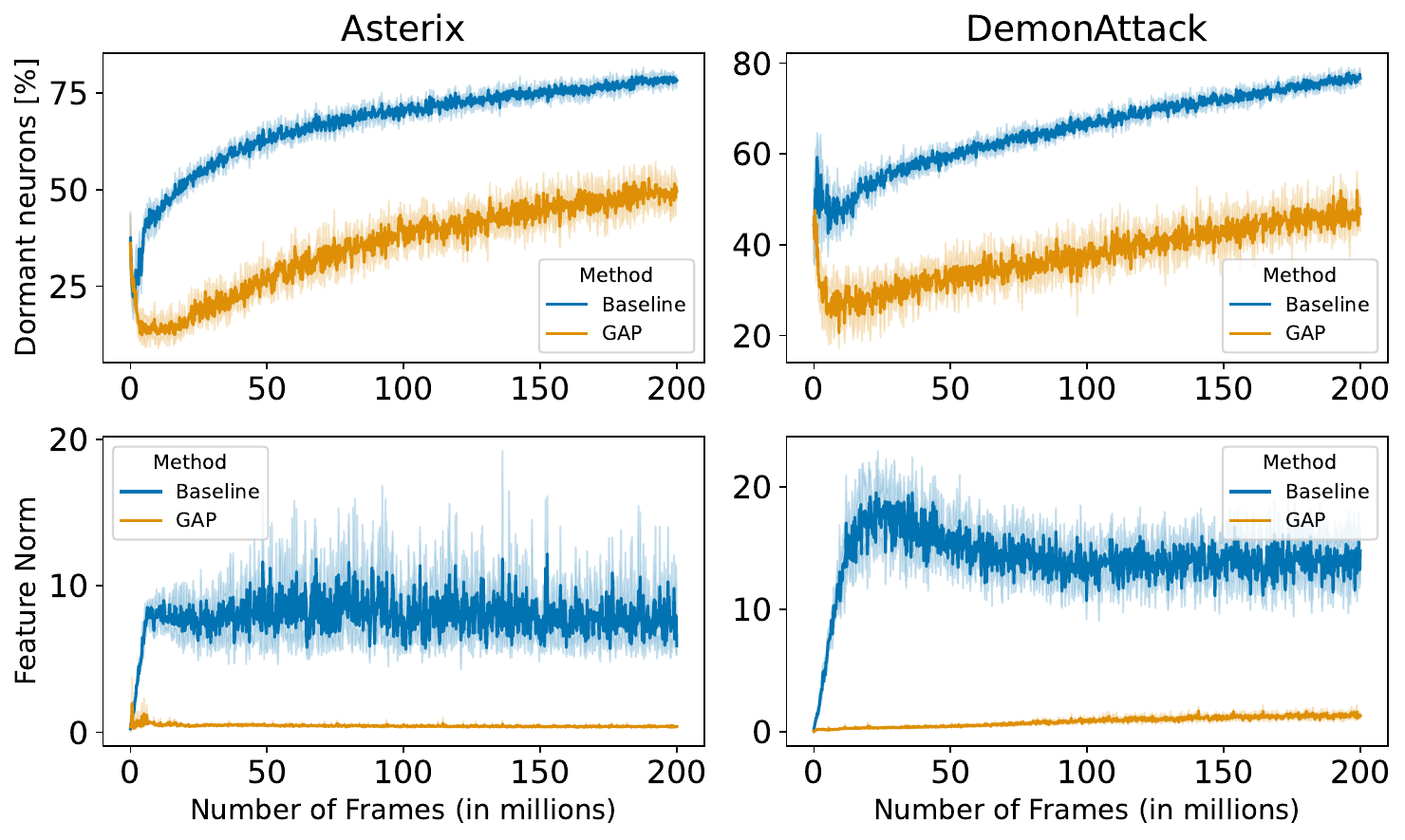}
\caption{Scaled networks with GAP exhibit less dormant neurons than the baseline and have lower feature norm.}
\label{fig:dormant_gap}
\end{minipage}
  \hspace{0.3cm}
  \begin{minipage}{0.43\linewidth}
    \centering
\includegraphics[width=0.7\columnwidth]{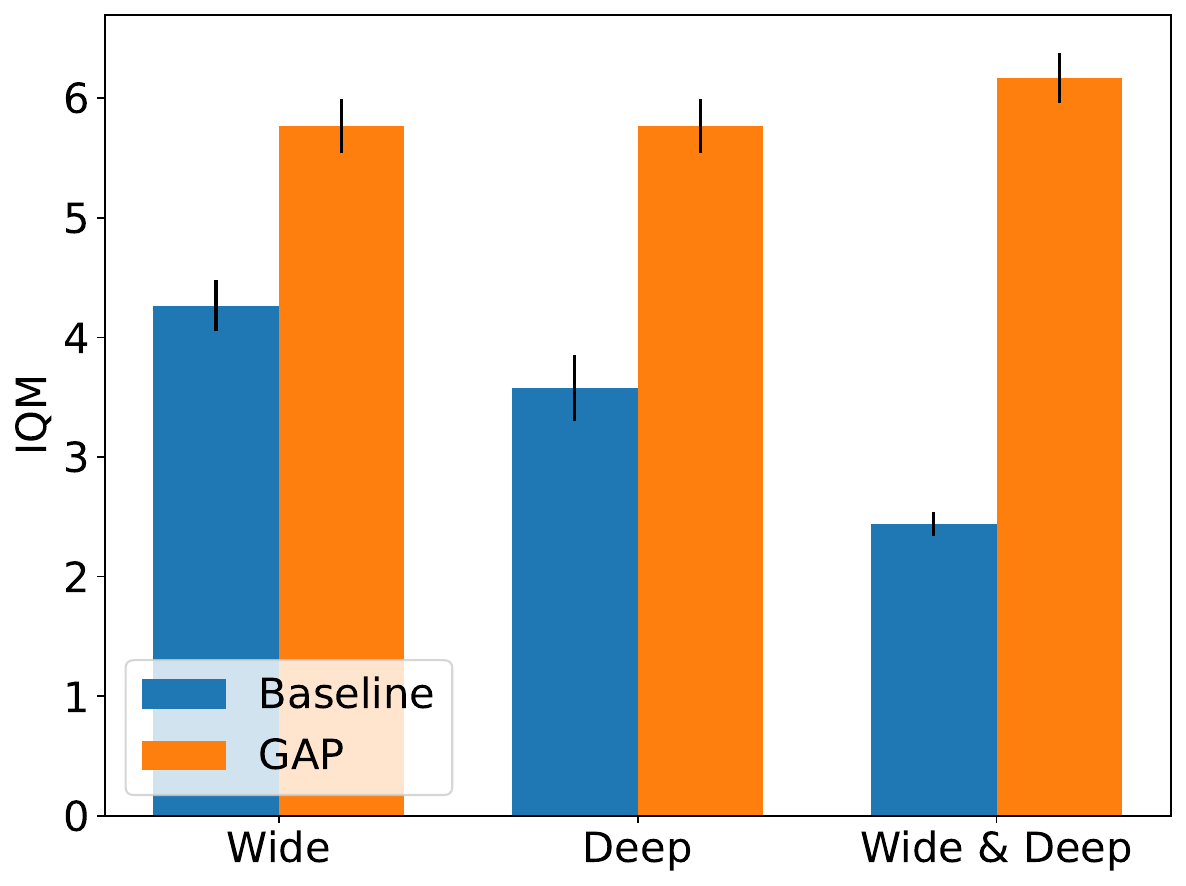}
    \caption{Impact of scaling network depth and width. Increasing depth hurts performance for the baseline networks, and scaling both width and depth makes it even worse. GAP, however, unlocks scaling even for increased network depth, leading to significant performance gains.}
    \label{fig:wide_deep}
   \end{minipage}   
\end{figure}

\begin{figure}[t]
  \centering
  \includegraphics[width=\textwidth]{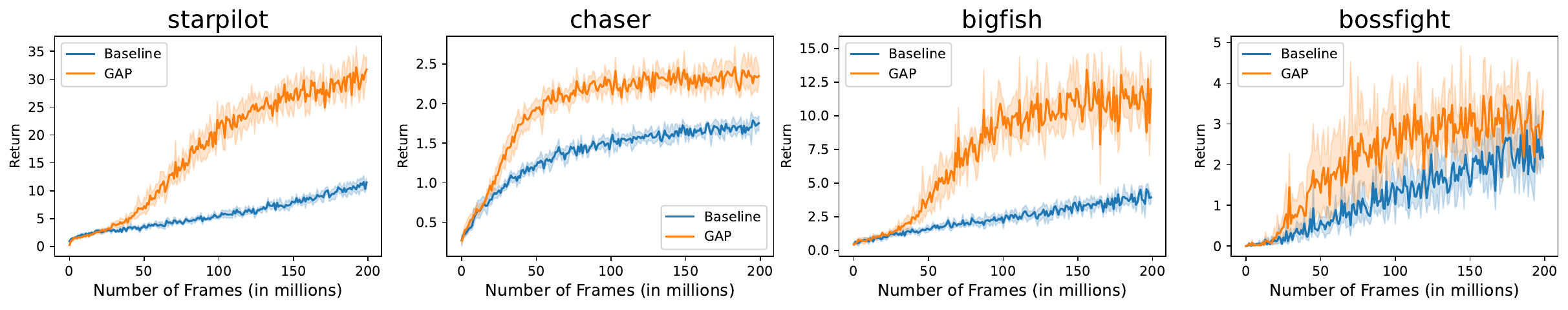}
    \includegraphics[width=\textwidth]{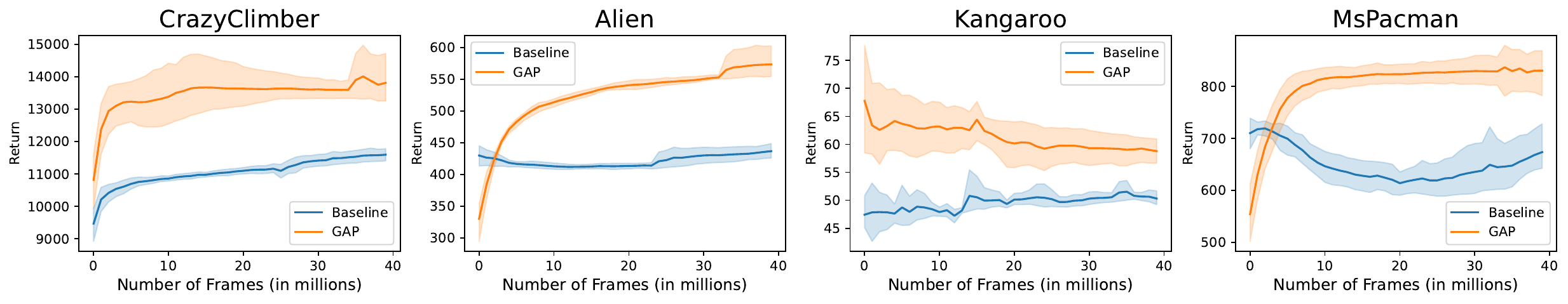}
    \includegraphics[width=\textwidth]{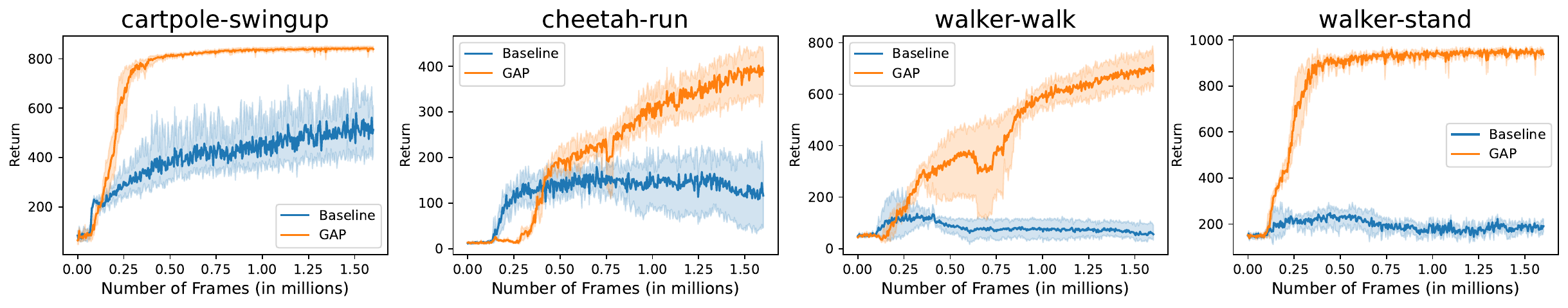}
  \caption{Performance for Rainbow on Procgen \citep{cobbe2019procgen} \textbf{(top)}, DER on Atari100K \citep{Kaiser2020Model} \textbf{(middle)}, and SAC on DMC \citep{tassa2018deepmind} \textbf{(bottom)}. GAP leads to performance gains for scaled networks in diverse domains.}
  \label{fig:other_domains}
\end{figure}

Having identified low-density and well-structured bottlenecks as the main factor enabling scaling networks, we demonstrate a \textit{simple} alternative to the more sophisticated techniques recently explored: Global Average Pooling (GAP) \citep{lin2013network}. We demonstrate its efficacy across wide range of settings and aspects, including: (1) \textit{effectiveness} in network scaling under different scales, architectures, and in sample-efficient regime with high replay ratios (\autoref{fig:GAP}); (2) \textit{computational efficiency} (\autoref{fig:performancevscomputation}); (3) improved training \textit{stability} and feature learning (\autoref{fig:dormant_gap}); (4) unlocking \textit{width and depth} scaling (\autoref{fig:wide_deep}); and (5) \textit{generalized gains} across various domains (\autoref{fig:other_domains}).   

\paragraph{Simple} The output feature maps of $\phi$, denoted as $F \in \mathbb{R}^{H \times W \times C}$, are processed by average pooling. For each feature map ($F^c$), GAP computes the average over its spatial dimensions, resulting in the output ($\textbf{g} \in \mathbb{R}^C$), which is then fed to the fully connected layers $\psi$ (see \autoref{fig:rlArchitecture} for an illustration):   

\begin{equation}
g^c = \frac{1}{H \times W} \sum_{i=1}^{H} \sum_{j=1}^{W} F^c_{ij}.
\end{equation}

\paragraph{Effective} We evaluate the impact of this architectural change on various settings. \textit{Scale}: We assess the performance of GAP across different network scales. The left plot of \autoref{fig:GAP} demonstrates that this simple architectural change unlocks scaling in RL networks, leading to significant performance improvements across different scales. \textit{Sample-efficient regime}: increasing the number of gradient updates per environment interaction (replay ratio) is favorable for sample-efficiency. Yet, higher replay ratios often hurt performance \citep{nikishin2022primacy}. We study the performance of the scaled network across varying the replay ratio values (default $=$ 0.25). 
We find that even in this challenging setting, GAP has very strong performance compared to the baseline of the same network size, yielding more sample-efficient agents, as shown in the center left plot of \autoref{fig:GAP}. \textit{Varying architecture}: we evaluate the effect of GAP on a different architecture for $\phi$: the original CNN architecture used in \citep{mnih2015humanlevel}. The right center plot in  \autoref{fig:GAP} shows that GAP can also provide performance gain to this architecture. \textit{Full suite:} we assess the generalization of our findings beyond the $20$ games used for most results and evaluate on the full set of 60 games of Atari. The rightmost plot of \autoref{fig:GAP} confirms that GAP consistently improves performance on the full suite.

\paragraph{Efficient} In \autoref{fig:performancevscomputation} we report the total number of GPU hours per game for every scale, to compare the computational cost of the different methods. While scaling the baseline increases computational costs and degrades the performance, applying GAP to the encoder's output reduces the effective density of $\psi$ (and hence computational cost), while yielding performance gains. Despite having a similar effective density to SoftMoE, GAP's inherent simplicity makes it more efficient by avoiding the extra computations associated with SoftMoE's token construction and post-MoE projection. The high runtime observed in sparse methods, despite their low effective density, stems from the fact that sparsity is only simulated with parameter masking  \citep{hoefler2021sparsity,mocanu2018scalable,evci2020rigging}.

\paragraph{Stable training dynamics} \autoref{fig:Saliencymap}~(right) displays the saliency maps when training with GAP, where the network seems to be attending to the most important and relevant areas of the input. We further examine the dormant neurons \citep{sokar2023dormant} and the norm of the features in \autoref{fig:dormant_gap}. We observe that the network exhibits fewer dormant neurons than the baseline and lower feature norms, suggesting improved plasticity and training stability.

\paragraph{Unlocking width-depth scaling} We demonstrate the effectiveness of GAP in scaling network width which is the focus of previous works \citep{ceron2024mixtures,ceron2024pruned,sokar2025dont}. Beyond this, we also explore how increasing $\psi$'s depth, or both width and depth, impacts the performance of RL networks. Specifically, we add extra fully connected layer in $\psi$ and evaluate the performance with both unscaled and scaled width of $\psi$. More analysis with varying depth are included in the appendix. We present the results in \autoref{fig:wide_deep}. We find that the baseline's performance drops with increased  depth, worsening significantly when scaling both dimensions. In contrast, interestingly GAP maintains strong performance across all scaling dimensions, confirming the impact of the representation learned by the bottleneck in the overall performance of the network.   
\begin{figure}[b]
  \centering
\includegraphics[width=0.94\textwidth]{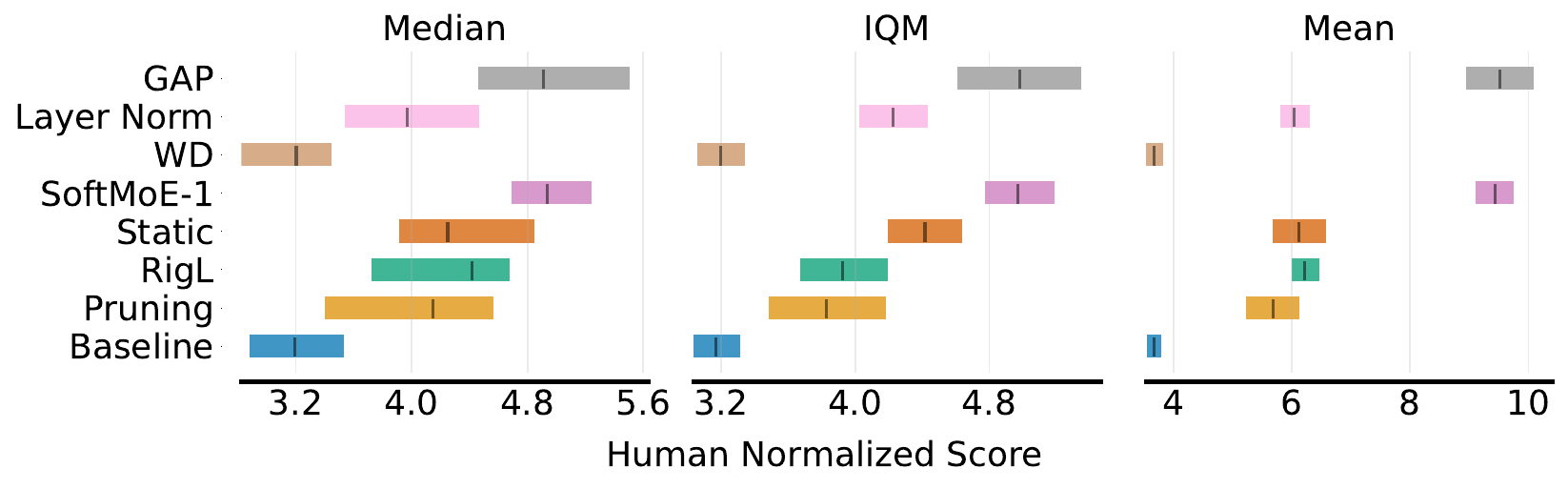}
  \caption{Comparison against architectural and algorithmic techniques for scaling RL networks. We report Median, IQM, and Mean scores \citep{agarwal2021deep} at 100M environment steps. GAP presents a notably simpler alternative to the baseline approaches, while achieving the best performance.}
  \label{fig:comparsion-against-other}
\end{figure}

\paragraph{Comparison against other methods} We compare the performance of GAP against various architectural techniques including pruning \citep{ceron2024pruned}, static and dynamic sparsity (RigL) \citep{graesser2022state}, and SoftMoE-1 \citep{sokar2025dont}. Moreover, we include a comparison against algorithmic methods including weight decay \citep{sokar2023dormant, ceron2024pruned}, and layer normalization \citep{nauman2024bigger}. \autoref{fig:comparsion-against-other} shows that GAP outperforms all baseline methods and achieves performance comparable to SoftMoE, while being more efficient and simpler. Although GAP is a more aggressive compression technique than granular approaches like per-conv tokens in SoftMoEs \citep{ceron2024mixtures}, which could lead to a loss of spatial structure, it still preserves spatial information within each feature map more effectively than a flattening operation.

\paragraph{Generalization to other domains} To further assess the broad applicability of our findings, we extended our evaluation to a diverse set of domains and agents. We assessed the Rainbow agent on Procgen \citep{cobbe2019procgen}, the DER agent \citep{van2019use} in the data-efficient  Atari100K setting \citep{Kaiser2020Model}, and the SAC agent \citep{haarnoja2018soft} on the continuous DeepMind Control (DMC) suite \citep{tassa2018deepmind}. For SAC, we follow the CNN architecture presented in \citep{yarats2021improving} and scale the embedding layer of the actor and critic by 8. \autoref{fig:other_domains} illustrates that these experiments align with our main results, confirming that GAP provides a consistent and significant performance benefit for scaled networks across these distinct benchmarks.

\section{Related Work}
\label{sec:related_work}

Several works have shown that scaling RL network causes substantial performance degradation due to training instabilities exhibited by the network \citep{hessel2018rainbow,bjorck2021towards,ceron2024mixtures}. Although the precise causes of these issues remain unclear, several approaches aim to mitigate them. We categorize these methods to architectural methods that alter the standard network architecture and algorithmic methods.  

\paragraph{Scaling through architectural changes} 
Recent works have investigated architectural modifications to improve the scaling of RL networks. \citet{ceron2024mixtures} integrated a Mixture-of-Experts (MoE) after the encoder in single-task RL networks. Their results across multiple domains highlight the effectiveness of this approach for scaling RL, with SoftMoE \citep{puigcerver2024from} outperforming traditional MoE \cite{shazeer2017outrageously}. This MoE approach was later extended by \citet{willi2024mixture}, who demonstrated its applicability in the multi-task setting. Relatedly, \citet{sokar2025dont} provided insights into why such methods might succeed, showing that replacing the standard flattened representation from the encoder with a tokenized representation that preserves spatial structure significantly improves the performance of scaled networks. Another line of research studies replacing dense parameters by sparse ones in both online \citep{graesser2022state,tanrlx2,sokar2022dynamic} and offline RL \citep{arnob2021single}. This approach has demonstrated its effectiveness in increasing the learning speed and performance of RL agents. Network sparsity is achieved either by starting with a dense network and progressively pruning weights, or by initializing with a sparse network and maintaining a consistent sparsity level throughout training. In the latter case, the sparse structure can be kept static or optimized dynamically during training using methods like SET \citep{mocanu2018scalable} and RigL \citep{evci2020rigging}. Recently, \citet{ceron2024pruned} showed that gradual magnitude pruning helps in scaling RL networks, leading to improved performance. Concurrent works have independently proposed similar approaches to GAP \citep{trumppimpoola,kooi2025hadamaxencodingelevatingperformance}, which provides extra evidence for the efficacy of this method.    

\paragraph{Scaling through algorithmic changes} A primary goal in this line of research is maintaining training stability as networks grow. \citet{bjorck2021towards} show that using spectral normalization \citep{miyato2018spectral} helps to improve training stability and enable using large neural networks for actor-critic methods. \citet{farebrother2023protovalue} explore the usage of auxiliary tasks to learn scaled representations. \citet{farebrother2024stop} demonstrated that training value networks using classification with categorical cross-entropy, as opposed to regression, leads to better performance in scaled networks. Other regularization techniques have been shown to be critical. \citet{schwarzer23bbf} propose several tricks to enable scaling including weight decay, network reset \citep{nikishin2022primacy}, increased discount factor, among others. Similarily, \citet{nauman2024bigger} employ a combination of layer normalization \citep{ba2016layer}, weight decay, and weight reset.

\section{Conclusion}
\label{sec:conclusion}
Recent proposals for enabling scaling in deep reinforcement learning have relied on sophisticated architectural interventions, such as the use of mixtures-of-experts and sparse training techniques. We demonstrated that these methods are indirectly targeting the {\bf bottleneck} connecting the encoder $\phi$ and dense layers $\psi$, in a standard deep RL network. A consequence of our analyses is that directly targeting this bottleneck, for instance with global average pooling, can achieve the same (or higher) performance gains.

Our work highlights the importance of better understanding the training dynamics of neural networks in the context of reinforcement learning. The fact that a simple technique like global average pooling can outperform existing literature suggests that architecture design is ripe for exploration. There is also a likely connection of our findings with works exploring representation learning for RL \citep{castro2021mico,kemertas2021towards,zhang2021learning,farebrother2023pvn}, given that these generally target the output of $\phi$; indeed, most of these methods aim to {\em structure} the outputs of $\phi$ so as to improve the generalizability and efficiency of the networks. It would be valuable to investigate whether approaches like GAP are complementary with (and ideally help enhance) more sophisticated representation learning approaches.

\paragraph{Limitations} Although our broad set of results suggest our findings are quite general, our investigation has focused on pixel-based environments where there is a clear bottleneck, or separation between $\phi$ (the convolutional layers) and $\psi$ (the fully connected layers). It is not clear whether our findings extend to non-pixel based environments or architectures where there isn't a clear bottleneck, but would be an interesting line of future work.

\section*{Acknowledgements}
The authors would like to thank Gheorghe Comanici, Joao Madeira Araujo, Karolina Dziugaite, Doina Precup, and the rest of the Google DeepMind team, as well as Roger Creus Castanyer and Johan Obando-Ceron, for valuable feedback on this work.

\newpage

\bibliographystyle{plainnat}
\bibliography{main}



\newpage

\section*{NeurIPS Paper Checklist}

\begin{enumerate}

\item {\bf Claims}
    \item[] Question: Do the main claims made in the abstract and introduction accurately reflect the paper's contributions and scope?
    \item[] Answer: \answerYes{} 
    \item[] Justification: The paper aims at understanding the challenges in network scaling in RL and underlying reasons behind the success of existing algorithms. The abstract and introduction include this.  

\item {\bf Limitations}
    \item[] Question: Does the paper discuss the limitations of the work performed by the authors?
    \item[] Answer: \answerYes{} 
    \item[] Justification: We discussed the limitations in Section \ref{sec:conclusion}.

\item {\bf Theory assumptions and proofs}
    \item[] Question: For each theoretical result, does the paper provide the full set of assumptions and a complete (and correct) proof?
    \item[] Answer: \answerNA{} 
    \item[] Justification: The paper does not include theoretical results. 

    \item {\bf Experimental result reproducibility}
    \item[] Question: Does the paper fully disclose all the information needed to reproduce the main experimental results of the paper to the extent that it affects the main claims and/or conclusions of the paper (regardless of whether the code and data are provided or not)?
    \item[] Answer: \answerYes{} 
    \item[] Justification: We provide all experimental details in Section \ref{sec:experimental_setup} and the appendix.

\item {\bf Open access to data and code}
    \item[] Question: Does the paper provide open access to the data and code, with sufficient instructions to faithfully reproduce the main experimental results, as described in supplemental material?
    \item[] Answer: \answerYes{} 
    \item[] Justification: The paper only uses already available open-source code, and provides sufficient instructions to faithfully reproduce.

\item {\bf Experimental setting/details}
    \item[] Question: Does the paper specify all the training and test details (e.g., data splits, hyperparameters, how they were chosen, type of optimizer, etc.) necessary to understand the results?
    \item[] Answer: \answerYes{} 
    \item[] Justification: We provide all experimental details in Section \ref{sec:experimental_setup}.

\item {\bf Experiment statistical significance}
    \item[] Question: Does the paper report error bars suitably and correctly defined or other appropriate information about the statistical significance of the experiments?
    \item[] Answer: \answerYes{} 
    \item[] Justification: We run each experiment using 5 seeds. Error bars represent 95\% stratified bootstrap confidence intervals. 

\item {\bf Experiments compute resources}
    \item[] Question: For each experiment, does the paper provide sufficient information on the computer resources (type of compute workers, memory, time of execution) needed to reproduce the experiments?
    \item[] Answer: \answerYes{} 
    \item[] Justification: We provide the compute resources in Section \ref{sec:experimental_setup}.
    
\item {\bf Code of ethics}
    \item[] Question: Does the research conducted in the paper conform, in every respect, with the NeurIPS Code of Ethics \url{https://neurips.cc/public/EthicsGuidelines}?
    \item[] Answer: \answerYes{} 
    \item[] Justification: The work conform with the NeurIPS Code of Ethics. 

\item {\bf Broader impacts}
    \item[] Question: Does the paper discuss both potential positive societal impacts and negative societal impacts of the work performed?
    \item[] Answer: \answerYes{} 
    \item[] Justification: We discuss the broader impacts in the appendix.
    
\item {\bf Safeguards}
    \item[] Question: Does the paper describe safeguards that have been put in place for responsible release of data or models that have a high risk for misuse (e.g., pretrained language models, image generators, or scraped datasets)?
    \item[] Answer: \answerNA{} 
    \item[] Justification: The paper poses no such risks.

\item {\bf Licenses for existing assets}
    \item[] Question: Are the creators or original owners of assets (e.g., code, data, models), used in the paper, properly credited and are the license and terms of use explicitly mentioned and properly respected?
    \item[] Answer: \answerYes{} 
    \item[] Justification: We include the license of used libraries.

\item {\bf New assets}
    \item[] Question: Are new assets introduced in the paper well documented and is the documentation provided alongside the assets?
    \item[] Answer: \answerNA{} 
    \item[] Justification: The paper does not release new assets.

\item {\bf Crowdsourcing and research with human subjects}
    \item[] Question: For crowdsourcing experiments and research with human subjects, does the paper include the full text of instructions given to participants and screenshots, if applicable, as well as details about compensation (if any)? 
    \item[] Answer: \answerNA{} 
    \item[] Justification: The paper does not involve crowdsourcing nor research with human subjects.

\item {\bf Institutional review board (IRB) approvals or equivalent for research with human subjects}
    \item[] Question: Does the paper describe potential risks incurred by study participants, whether such risks were disclosed to the subjects, and whether Institutional Review Board (IRB) approvals (or an equivalent approval/review based on the requirements of your country or institution) were obtained?
    \item[] Answer: \answerNA{} 
    \item[] Justification: The paper does not involve crowdsourcing nor research with human subjects.

\item {\bf Declaration of LLM usage}
    \item[] Question: Does the paper describe the usage of LLMs if it is an important, original, or non-standard component of the core methods in this research? Note that if the LLM is used only for writing, editing, or formatting purposes and does not impact the core methodology, scientific rigorousness, or originality of the research, declaration is not required.
    \item[] Answer: \answerNA{} 
    \item[] Justification: The paper does not include LLMs.

\end{enumerate}
\newpage
\appendix
\section{Broader impacts}
This paper studies the challenges in scaling networks in pixel-based deep reinforcement learning. We present a simple, effective alternative to current sophisticated approaches by identifying the effective network representation. The approach has a positive impact by requiring no hyperparameters, simplifying implementation and showing robust performance across various scenarios. While this research aims to advance RL agent capabilities without direct negative impact, we urge careful consideration of potential implications when building upon this work.

\section{Experimental Details}

\begin{table}[!h]
 \centering
  \caption{Hyper-parameters for Rainbow and DER agents.}
  \label{tbl:defaultvalues}
 \begin{tabular}{@{} cccc @{}}
    \toprule
    & \multicolumn{3}{c}{Atari}\\
  & Hyper-parameter & Rainbow & DER\\
    \midrule
 \multirow{4}{*}{Training}   & Adam's ($\epsilon$) & 1.5e-4 & 0.00015\\
     &Adam's learning rate & 6.25e-5 & 0.0001\\
     &Batch Size & 32 & 32\\
     &Weight Decay  & 0 & 0\\
     \hline
    \multirow{2}{*}{Architecture} & Activation Function  & ReLU & ReLU\\
     & Fully connected layer Width & 512 & 512\\
     \hline     
   \multirow{8}{*}{Algorithm} & Replay Capacity & 1000000& 1000000 \\
     & Minimum Replay History & 20000& 1600\\
     & Number of Atoms  & 51 & 51 \\
     & Reward Clipping  & True & True\\
     & Update Horizon  & 3 & 10\\
     & Update Period  & 4 &  1\\
     & Discount Factor  & 0.99 & 0.99 \\
     & Exploration $\epsilon$ & 0.01 & 0.01\\
     & Sticky Actions & True & False\\
     \bottomrule
  \end{tabular}
\end{table}

\paragraph{Hyperparameter details} We use the default hyperparameters for all the studied algorithms. We present the values of these parameters in \autoref{tbl:defaultvalues}. For the dormant neuron analysis, we use a dormancy threshold of $0.001$. For the feature learning analysis (\autoref{fig:feature_learning}), we increase the depth of the encoder by adding two ResNet blocks. 

\paragraph{Atari Games \citep{bellemare2013arcade}} We use the set of 20 games used in \citet{sokar2025dont,ceron2024mixtures} for direct comparison. The set has the following games: Asterix, SpaceInvaders, Breakout, Pong, Qbert, DemonAttack, Seaquest, WizardOfWor, RoadRunner, BeamRider, Frostbite, CrazyClimber, Assault, Krull, Boxing, Jamesbond, Kangaroo, UpNDown, Gopher, and Hero. This set is used in most of our analysis, nevertheless we provide our main results on the full suite of $60$ games. 

\paragraph{Atari100K Games \citep{Kaiser2020Model}} We test on the $26$ games of this benchmark. It includes the following games: Alien, Amidar, Assault, Asterix, BankHeist, BattleZone, Boxing, Breakout, ChopperCommand, CrazyClimber, DemonAttack, Freeway, Frostbite, Gopher, Hero, Jamesbond, Kangaroo, Krull, KungFuMaster, MsPacman, Pong, PrivateEye, Qbert, RoadRunner, Seaquest, UpNDown.

\section{Extra Experiments}
\begin{figure}[t]
  \centering
  \includegraphics[width=0.45\textwidth]{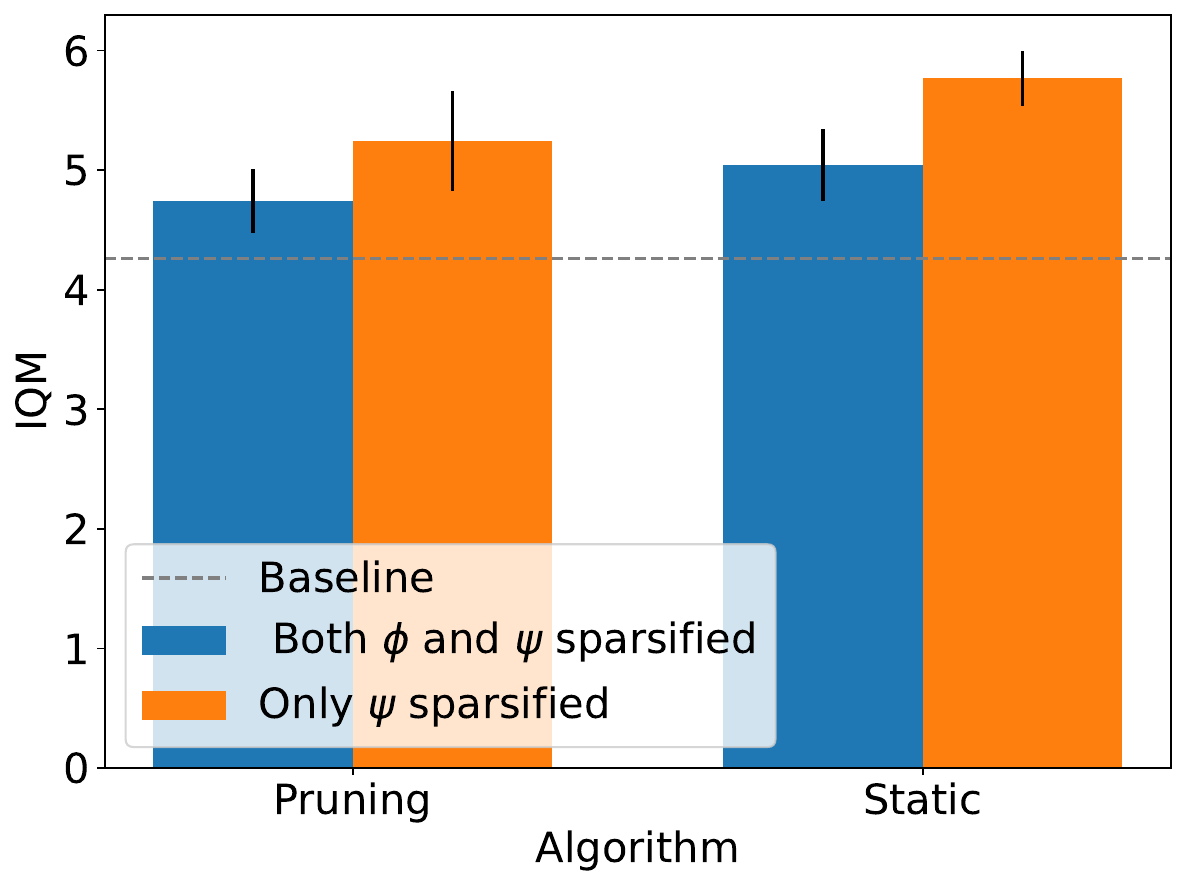}
  \vspace{0.5cm}
    \includegraphics[width=0.45\textwidth]{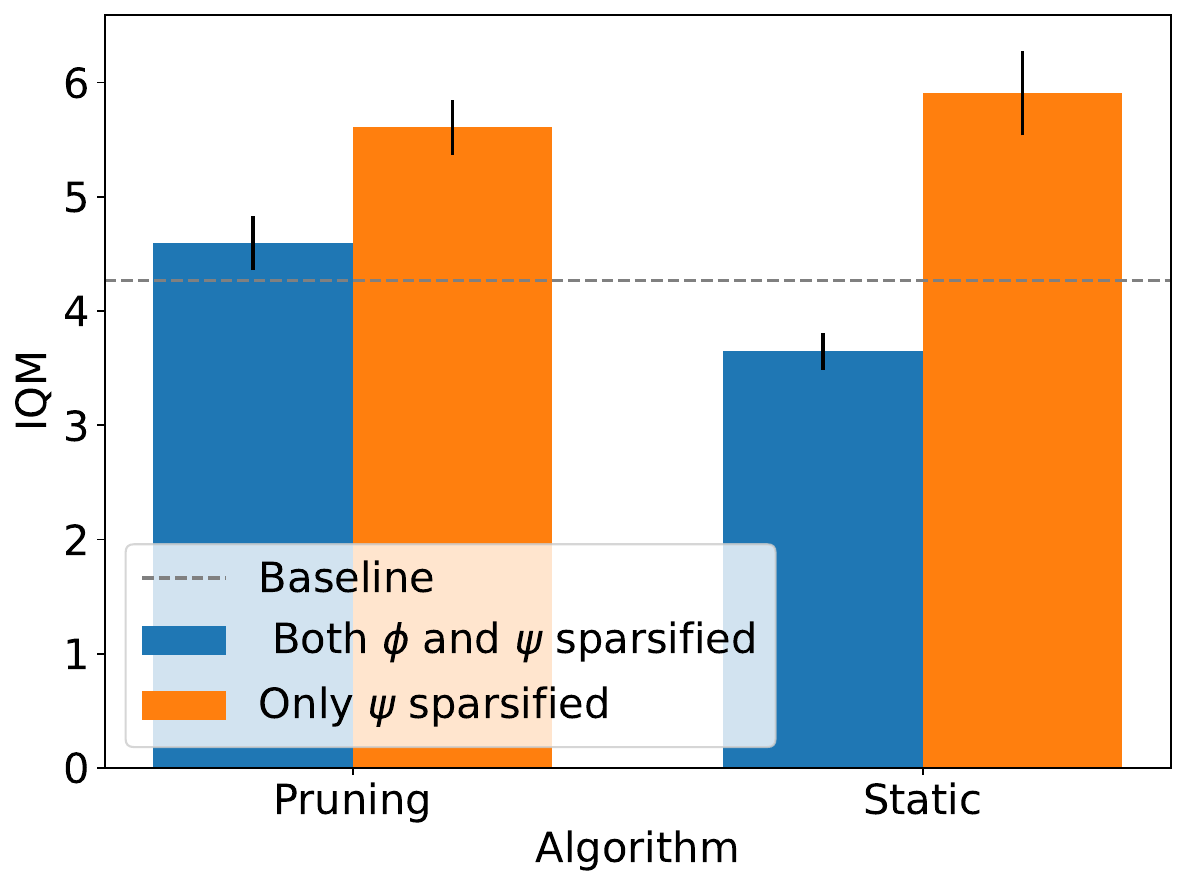}
  \caption{Sparsification of $\psi$ yields better performance than sparsifying $\phi$ and $\psi$ for 80\% sparsity \textbf{(left)} and 95\% sparsity \textbf{(right)}.}
  \label{fig:sparsity_all_vs_dense_80_95}
\end{figure}

\paragraph{Sparse methods address the bottleneck} Extending our analysis in \autoref{sec:analyses}, we validate our hypothesis on various sparsity levels. Consistent with our main results, sparsification of only $\psi$ yields better performance than sparsifying $\phi$ and $\psi$ across all sparsity levels as shown in \autoref{fig:sparsity_all_vs_dense_80_95}.

\begin{figure}
  \centering
  \includegraphics[width=\columnwidth]{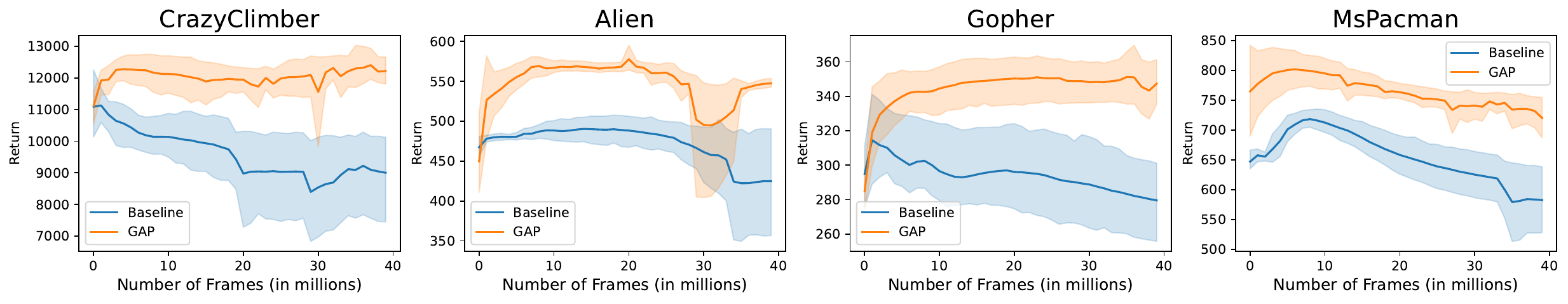}
    \includegraphics[width=\columnwidth]{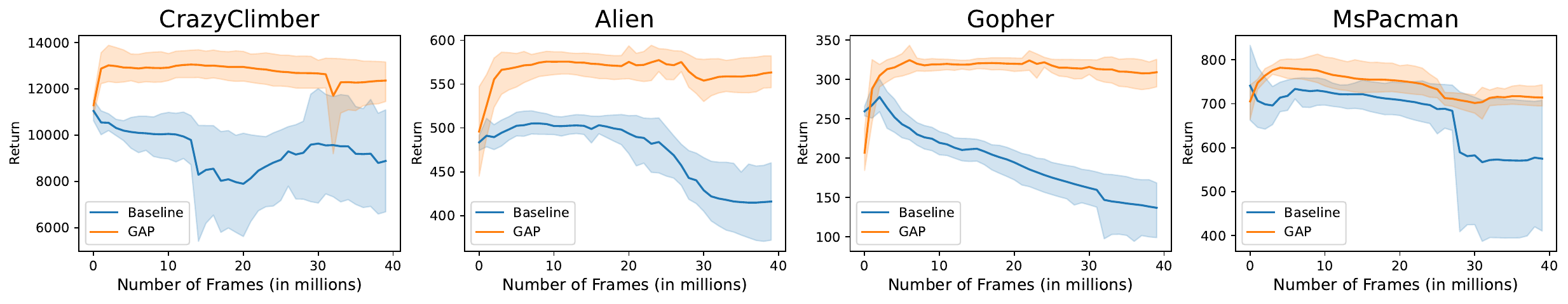}
  \caption{Performance for DrQ \textbf{(top)} and DrQ($\epsilon$) \textbf{(bottom)} \citep{yarats2021image,agarwal2021deep} on Atari100K.}
  \label{fig:drqdrqeps}
\end{figure}

\paragraph{More agents} We evaluate two more algorithms in the sample-efficient regime: DrQ and DrQ($\epsilon$) \citep{yarats2021image,agarwal2021deep} on Atari100K. Consistent with our main results, GAP improves performance of scaled networks as shown in \autoref{fig:drqdrqeps}.

\begin{figure}[t]
  \centering
    \begin{minipage}{0.3\linewidth}
        \centering
  \includegraphics[width=\columnwidth]{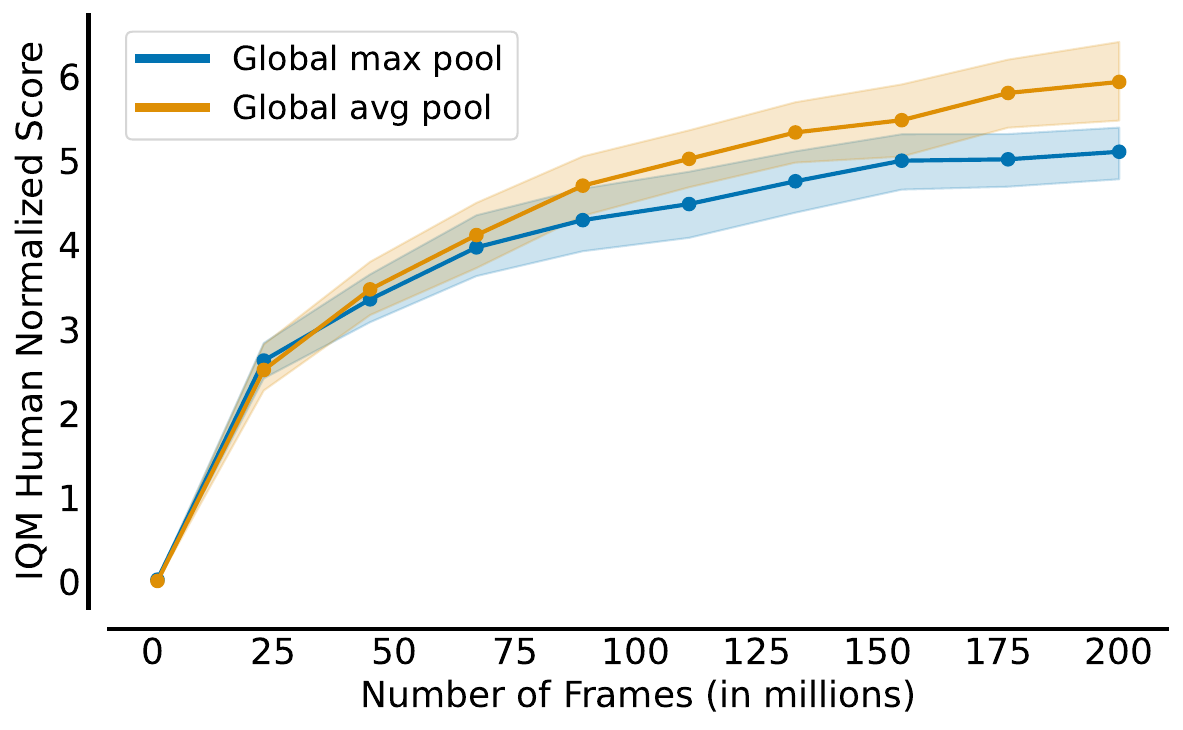}
  \caption{Comparison between GAP and global max pooling for Rainbow.}
  \label{fig:pooling_types}
\end{minipage}
  \hspace{0.3cm}
  \begin{minipage}{0.3\linewidth}
    \centering
  \includegraphics[width=\columnwidth]{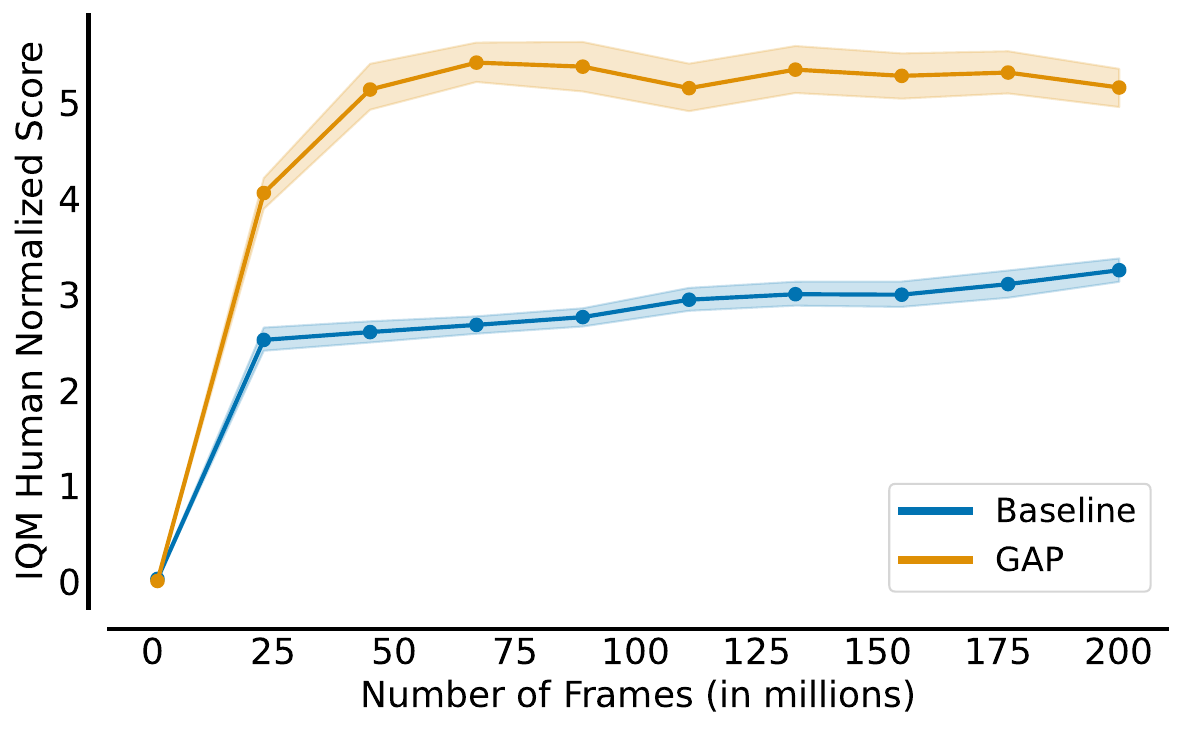}
  \caption{Effect of increasing the width of $\phi$. GAP improves performance over the scaled baseline.}
  \label{fig:wider_phi_GAP_baseline}
    \end{minipage}  
    \hspace{0.3cm}
      \begin{minipage}{0.3\linewidth}
    \centering
  \includegraphics[width=\columnwidth]{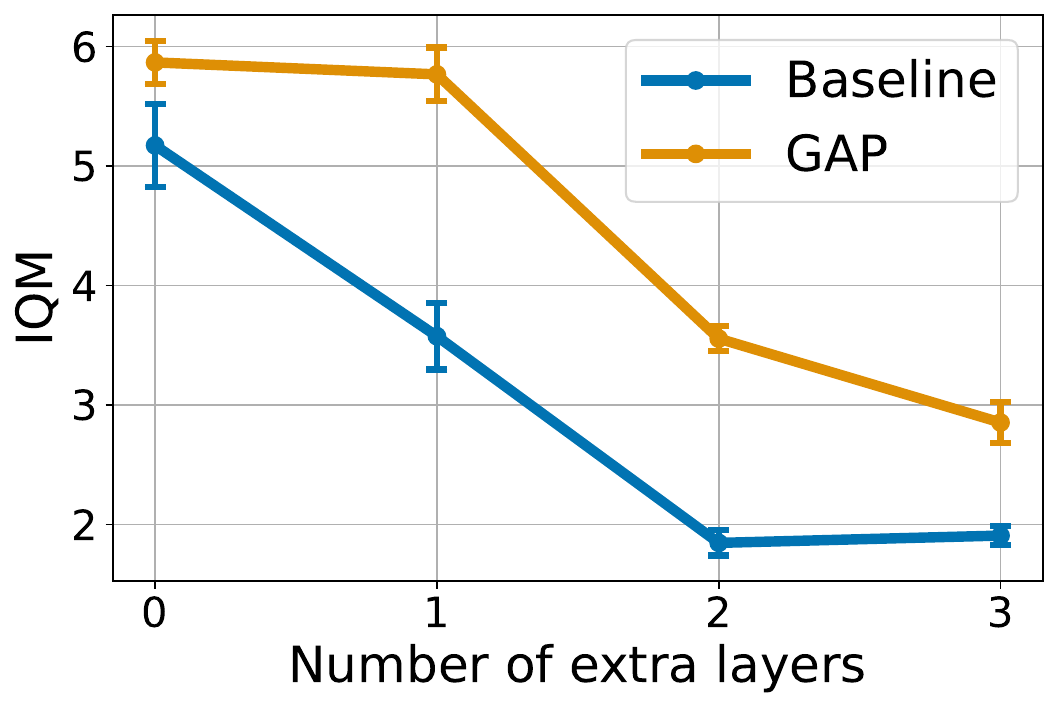}
  \caption{Effect of increasing the depth of $\psi$. GAP achieves better performance than the corresponding baseline with the same network size.}
  \label{fig:deeper_GAP_baseline}
   \end{minipage}   
\end{figure}

\paragraph{Comparison against max pooling}
We compare the performance of global average pooling with global max pooling for Rainbow on the set of 20 games used in our main results. We find the GAP outperforms max pooling as demonstrated in \autoref{fig:pooling_types}. We hypothesize this is due to GAP's ability to retain more comprehensive information through its averaging operation. 

\paragraph{Encoder width scaling} We perform additional experiment analyzing the effect of scaling the width of layers in $\phi$ by a factor of 4, while maintaining $\psi$ unscaled. \autoref{fig:wider_phi_GAP_baseline} shows that GAP significantly improves performance. 

\paragraph{Deeper networks} While our main focus in this work is scaling the width of the network, we also performed some preliminary analysis on scaling the depth of the fully connected layers ($\psi$), exploring architectures with 1, 2, and 3 additional layers. As shown in \autoref{fig:deeper_GAP_baseline}, increasing the depth degrades the performance of the agent. Increasingly, GAP improves performance over the corresponding baseline network of the same size across varying depth.  

\begin{figure}
  \centering
  \includegraphics[width=0.8\textwidth]{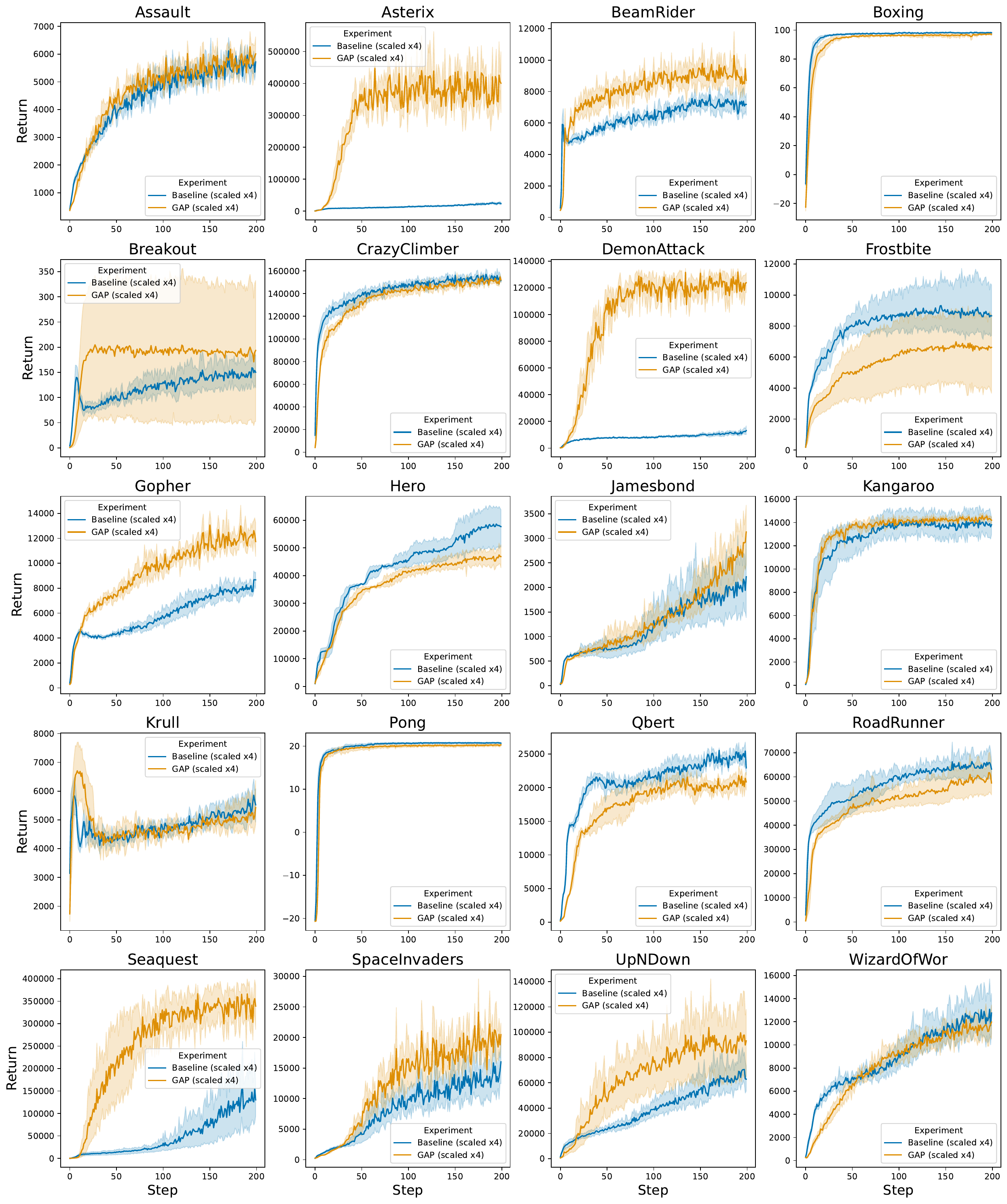}
  \caption{Performance during training on the 20 games of the Atrai benchmark.}
  \label{fig:allgames_Atari}
\end{figure}

\begin{figure}
  \centering
  \includegraphics[width=0.8\textwidth]{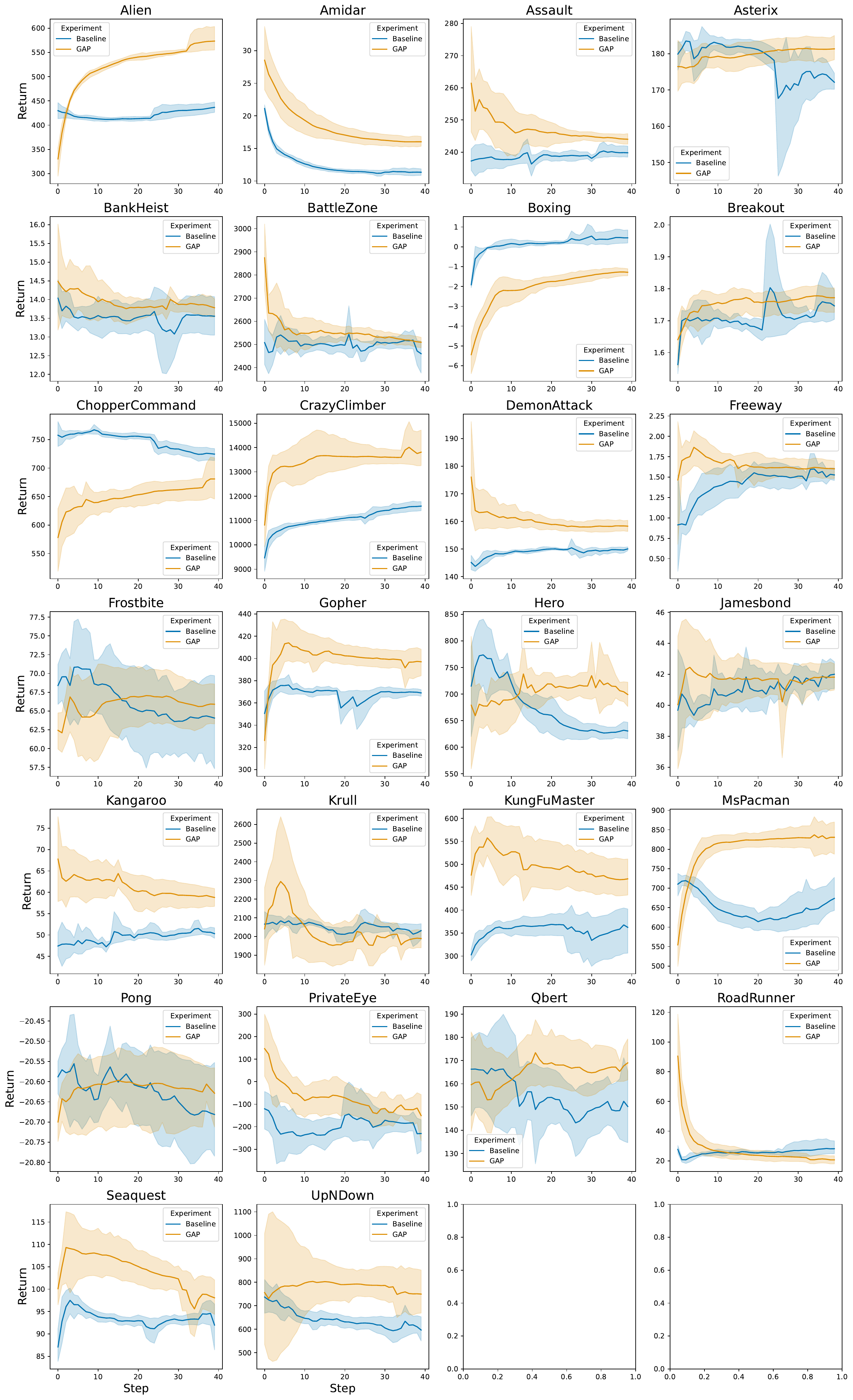}
  \caption{Performance during training on the 26 games of the Atrai100k benchmark.}
  \label{fig:allgames_Atari100k}
\end{figure}

\paragraph{Performance throughout training}
\autoref{fig:allgames_Atari} and \autoref{fig:allgames_Atari100k} present the performance per game throughout training for Atari and Atari100K benchmarks, respectively.

\end{document}